\documentclass[graybox]{svmult}

\usepackage{mathptmx}       % selects Times Roman as basic font
\usepackage{helvet}         % selects Helvetica as sans-serif font
\usepackage{courier}        % selects Courier as typewriter font
\usepackage{type1cm}        % activate if the above 3 fonts are
\usepackage{makeidx}         % allows index generation
\usepackage{graphicx}        % standard LaTeX graphics tool
\usepackage{multicol}        % used for the two-column index
\usepackage[bottom]{footmisc}% places footnotes at page bottom

\usepackage{amsfonts}
\usepackage{amsmath}
\usepackage{bm}
\usepackage{array}
\usepackage{subscript} % for \textsubscript
\usepackage{url}

\usepackage{algorithm,algpseudocode}
\usepackage{subcaption}
\makeatletter
\newcommand{\algmargin}{\the\ALG@thistlm}   
\makeatother
\algnewcommand{\parState}[1]{\State%
    \parbox[t]{\dimexpr\linewidth-\algmargin}{\strut\hangindent=\algorithmicindent \hangafter=1 #1\strut}}

\newif\ifMIT
\MITfalse
\def\secref#1{section~\ref{#1}}
\def\eqref#1{equation~\ref{#1}}
\def\1{\bm{1}}

\ifMIT
\def\gamma{{\boldmath{\vgamma}}}

\else

\def\vgamma{{\bm{\gamma}}}

\fi

\ifMIT

\else

\fi

\ifMIT
\else
\DeclareMathOperator*{\argmax}{arg\,max}

\fi
\DeclareMathOperator{\sign}{sign}

\makeindex             % used for the subject index
\begin{document}

\title*{Adversarial Attacks and Defences Competition}


% Comment next line for camera ready version of the chapter.
\titlerunning{This is a preprint of a Springer book chapter from the ``NIPS 2017 Competition Book''}

\author{Alexey Kurakin and Ian Goodfellow and Samy Bengio
% 1st place (all tracks) - team TsAIL
and Yinpeng Dong and Fangzhou Liao and Ming Liang and Tianyu Pang and Jun Zhu and Xiaolin Hu
% 2nd place defense - team iyswim
and Cihang Xie and Jianyu Wang and Zhishuai Zhang and Zhou Ren and Alan Yuille
% 2nd place non-targeted attack and targeted attack - team Sangxia
and Sangxia Huang
% 3rd place defense and 4th place targeted attack - team Anil Thomas
% 3rd place non-targeted attack - team Stanford & Sun
% 3rd place targeted attack - team FatFingers
and Yao Zhao and Yuzhe Zhao and Zhonglin Han and Junjiajia Long
% 4th place defense - team erko
and Yerkebulan Berdibekov
% 4th place non-targeted attack - team iwiwi
and Takuya Akiba and Seiya Tokui and Motoki Abe}

\authorrunning{Kurakin et al.}


\institute{Alexey Kurakin, Ian Goodfellow, Samy Bengio \at Google Brain
% 1st place (all tracks) - team TsAIL
\and Yinpeng Dong, Fangzhou Liao, Ming Liang, Tianyu Pang, Jun Zhu, Xiaolin Hu \at Department of Computer Science and Technology, Tsinghua University
% 2nd place defense - team iyswim
\and Cihang Xie, Zhishuai Zhang, Alan Yuille \at Department of Computer Science, The Johns Hopkins University
\and Jianyu Wang \at Baidu Research USA
\and Zhou Ren \at Snap Inc.
% 2nd place non-targeted attack and targeted attack - team Sangxia
\and Sangxia Huang \at Sony Mobile Communications, Lund, Sweden
% 3rd place defense and 4th place targeted attack - team Anil Thomas
% 3rd place non-targeted attack - team Stanford & Sun
% 3rd place targeted attack - team FatFingers
\and Yao Zhao \at Microsoft corp.
\and Yuzhe Zhao \at Dept of Computer Science, Yale Univerisity
\and Zhonglin Han \at Smule Inc.
\and Junjiajia Long \at Dept of Physics, Yale University
% 4th place defense - team erko
\and Yerkebulan Berdibekov \at Independent Scholar
% 4th place non-targeted attack - team iwiwi
\and Takuya Akiba, Seiya Tokui, Motoki Abe \at Preferred Networks, Inc.}
% other authors
%\and Other Authors \at Name, Address of Institute \email{name@email.address}}
\maketitle

\abstract{To accelerate research on adversarial examples and robustness of
machine learning classifiers, Google Brain organized a NIPS 2017 competition
that encouraged researchers to develop new methods to generate adversarial examples
as well as to develop new ways to defend against them.
In this chapter, we describe the structure and organization of the competition
and the solutions developed by several of the top-placing teams.
}

\section{Introduction}
\label{sec:adv_comp:intro}

% This section and subsections will contain introductory description
% about adversarial examples and why this problem is important:
%
% \begin{itemize}
%   \item What are adversarial examples
%   \item Why they are important
%   \item What are attack scenarios: black box attack and white box attack.
%   \item What are typical attacks (FGSM, PGD, C\&W and few others).
%   \item What are know ways to defend against adversarial examples.
% \end{itemize}

% introduction - what are adversarial examples and why they are important
Recent advances in machine learning and deep neural networks enabled researchers to solve multiple
important practical problems like image, video, text classification and others.

However most existing machine learning classifiers are highly vulnerable to adversarial
examples~\cite{biggio2013evasion,Szegedy-ICLR2014,Goodfellow-2015-adversarial,Papernot-2016-TransferabilityStudy}.
An adversarial example is a sample of input data which has been modified
very slightly in a way that is intended to cause a machine learning classifier
to misclassify it. In many cases, these modifications can be so subtle that a human
observer does not even notice the modification at all, yet the classifier still makes
a mistake.

Adversarial examples pose security concerns because they could be
used to perform an attack on machine learning systems, even if the adversary has
no access to the underlying model.

Moreover it was discovered~\cite{PhysicalAdversarialExamples,Sharif16AdvML}
that it is possible to perform adversarial attacks even on a machine learning system
which operates in physical world and perceives input through inaccurate sensors,
instead of reading precise digital data.

In the long run, machine learning and AI systems will become more powerful.
Machine learning security vulnerabilities similar to adversarial examples could
be used to compromise and control highly powerful AIs.
Thus, robustness to adversarial examples is an important part of the AI safety problem.

Research on adversarial attacks and defenses is difficult for many reasons.
One reason is that evaluation of proposed attacks or proposed defenses is
not straightforward.
Traditional machine learning, with an assumption of a training set and test
set that have been drawn i.i.d., is straightforward to evaluate by measuring
the loss on the test set.
For adversarial machine learning, defenders must contend with an open-ended
problem, in which an attacker will send inputs from an unknown distribution.
It is not sufficient to benchmark a defense against a single attack or even
a suite of attacks prepared ahead of time by the researcher proposing the
defense. Even if the defense performs well in such an experiment, it may
be defeated by a new attack that works in a way the defender did not anticipate.
Ideally, a defense would be provably sound, but machine learning in general
and deep neural networks in particular are difficult to analyze theoretically.
A competition therefore gives a useful intermediate form of evaluation:
a defense is pitted against attacks built by independent teams, with both
the defense team and the attack team incentivized to win.
While such an evaluation is not as conclusive as a theoretical proof, it
is a much better simulation of a real-life security scenario than an
evaluation of a defense carried out by the proposer of the defense.

In this report, we describe the NIPS 2017 competition on adversarial
attack and defense, including an overview of the key research
problems involving adversarial examples (\secref{sec:adv_comp:adv_examples}),
the structure and organization of
the competition (\secref{sec:adv_comp:competition}),
and several of the methods developed by the top-placing
competitors (\secref{sec:adv_comp:submissions}).

% maybe more formal definition

\section{Adversarial examples}
\label{sec:adv_comp:adv_examples}

Adversarial examples are inputs to machine learning models that have
been intentionally optimized to cause the model to make a mistake.
We call an input example a ``clean example'' if it is a naturally 
occurring example, such as a photograph from the ImageNet dataset.
If an adversary has modified an example with the intention of
causing it to be misclassified, we call it an ``adversarial example.''
Of course, the adversary may not necessarily succeed; a model
may still classify the adversarial example correctly.
We can measure the accuracy or the error rate of different models
on a particular set of adversarial examples.

\subsection{Common attack scenarios}

Scenarios of possible adversarial attacks can be categorized along different
dimensions.

First of all, attacks can be classified by the type of outcome the adversary
desires:

\begin{itemize}
\item \textbf{Non-targeted attack.} In this the case adversary's goal is to
  cause the classifier to predict any inccorect label.
  The specific incorrect label does not matter.
\item \textbf{Targeted attack.} In this case the adversary aims to change the
  classifier's prediction to some specific target class.
\end{itemize}

Second, attack scenarios can be classified by the amount of knowledge the
adversary has about the model:

\begin{itemize}
\item \textbf{White box.} In the white box scenario, the adversary has full
  knowledge of the model including model type, model architecture and
  values of all parameters and trainable weights.
\item \textbf{Black box with probing.} In this scenario, the adversary
  does not know very much about the model, but
  can probe or query the model, i.e. feed some inputs and observe outputs.
  There are many variants of this scenario---the adversary may know the architecture
    but not the parameters or the adversary may not even know the architecture,
    the adversary may be able to observe output probabilities for each class or
    the adversary may only be to observe the choice of the most likely class.
\item \textbf{Black box without probing} In the black box without probing scenario,
  the adversary
  has limited or no knowledge about the model under attack
  and is not allowed to probe or query the model while constructing adversarial examples.
  In this case, the attacker must construct adversarial examples that fool most machine
    learning models.
\end{itemize}

Third, attacks can be classifier by the way adversary can feed data into the model:

\begin{itemize}
\item \textbf{Digital attack.} In this case, the adversary has direct access to the
  actual data fed into the model. In other words, the adversary can choose
  specific {\tt float32} values as input for the model.
  In a real world setting, this might occur when an attacker uploads a PNG file
  to a web service, and intentionally designs the file to be read incorrectly.
  For example, spam content might be posted on social media, using adversarial
  perturbations of the image file to evade the spam detector.
\item \textbf{Physical attack.} In the case of an attack in the physical
  world, the adversary does not have direct access to the digital
  representation of provided to the model. Instead, the model is fed input
  obtained by sensors such as a camera or microphone. The adversary is able to
  place objects in the physical environment seen by the camera or produce
  sounds heard by the microphone. The exact digital representation obtained by
  the sensors will change based on factors like the camera angle, the distance
  to the microphone, ambient light or sound in the environment, etc.
  This means the attacker has less precise control over the input provided to
  the machine learning model.
\end{itemize}

\subsection{Attack methods}

Most of the attacks discussed in the literature are geared toward the white-box
digital case.

\subsubsection{White box digital attacks}

\runinhead{L-BFGS}.
One of the first methods to find adversarial examples for neural networks was proposed in~\cite{Szegedy-ICLR2014}.
The idea of this method is to solve the following optimization problem:

\begin{equation}
\left|x^{adv} - x\right|_2 \rightarrow \text{minimum}, \quad
\text{s.t.} \quad  f(x^{adv})=y_{target}, \quad x^{adv} \in [0, 1]^m
\end{equation}

The authors proposed to use the L-BFGS optimization method to solve this
problem, thus the name of the attack.

One of the main drawbacks of this method is that it is quite slow.
The method is not designed to counteract defenses such as reducing
the number of bits used to store each pixel.
Instead, the method is designed to find the smallest possible attack
perturbation. This means the method can sometimes be defeated merely
by degrading the image quality, for example, by rounding to an 8-bit
representation of each pixel.

\runinhead{Fast gradient sign method (FGSM).}
To test the idea that adversarial examples can be found using only a
linear approximation of the target model, the authors of ~\cite{Goodfellow-2015-adversarial}
introduced the {\em fast gradient sign method} (FGSM).

FGSM works by linearizing loss function in $L_{\infty}$ neighbourhood of a clean image and finds exact maximum of
linearized function using following closed-form equation:

\begin{equation}
x^{adv} = x + \epsilon \sign \bigl( \nabla_x J(x, y_{true})  \bigr)
\end{equation}

\iffalse  Ian commented the next part out just because it may be more detail than needed.
          OK to bring it back in, just seems like it could be longer than people want to read

Also this method provides only non-targeted attack. % There is a targeted version, but I guess its success rate is extremely low in a lot of cases? It worked on CIFAR for Ian

Another problem is that this method uses true labels, which may be not available to attacker.
Moreover the fact that method uses true labels may lead to ``label leaking'' phenomenon~\cite{Kurakin-AdversarialMlAtScale}
when it actually increases classifier accuracy of adversarially trained network, instead of confusing classifier.

Problem with true labels could be solved in several ways.
One way is to use network predictions $y_{pred} = \argmax f(x)$ instead of $y_{true}$.
Another alternative is to change the method to maximize probability $p(y_{non-true}|x)$ on some non-true class instead of maximizing value of 
loss function~\cite{Kurakin-PhysicalAdversarialExamples}, in such case it will lead to following formula:

\begin{equation}
x^{adv} = x - \epsilon \sign \bigl( \nabla_x J(x, y_{non-true})  \bigr)
\end{equation}

You can either pick $y_{non-true}$ randomly or choose least likely network prediction.
All of this variations of FGSM perform roughly similar~\cite{Kurakin-AdversarialMlAtScale}.
\fi

\runinhead{Iterative attacks}

The L-BFGS attack has a high success rate and high computational cost.
The FGSM attack has a low success rate (especially when the defender anticipates it)
and low computational cost.
A nice tradeoff can be achieved by running iterative optimization algorithms that
are specialized to reach a solution quickly, after a small number (e.g. 40) of iterations.

One strategy for designing optimization algorithms quickly is to take the FGSM (which can
often reach an acceptable solution in one very large step) and run it for several steps
but with a smaller step size. Because each FGSM step is designed to go all the way to
the edge of a small norm ball surrounding the starting point for the step, the method
makes rapid progress even when gradients are small.
This leads to the \textbf{Basic Iterative Method (BIM)} method introduced in~\cite{Kurakin-PhysicalAdversarialExamples}, also sometimes called \textbf{Iterative FGSM (I-FGSM)}:

\begin{equation}
x^{adv}_{0} = \bm{X}, \quad
x^{adv}_{N+1} = Clip_{X, \epsilon}\Bigl\{ \bm{X}^{adv}_{N} + \alpha \sign \bigl( \nabla_X J(\bm{X}^{adv}_{N}, y_{true})  \bigr) \Bigr\}
\end{equation}

The BIM can be easily made into a target attack, called the Iterative Target Class Method:

\begin{equation}
%y_{LL} = \argmin_{y} \bigl\{ p( y | \bm{X} ) \bigr\} \\
\bm{X}^{adv}_{0} = \bm{X}, \quad
\bm{X}^{adv}_{N+1} = Clip_{X, \epsilon}\left\{ \bm{X}^{adv}_{N} - \alpha \sign \left( \nabla_X J(\bm{X}^{adv}_{N}, y_{target})  \right) \right\}
\end{equation}

It was observed that with sufficient number of iterations this attack almost
always succeeds in hitting target class~\cite{Kurakin-PhysicalAdversarialExamples}.

\runinhead{Madry et. al's Attack}
\cite{MadryPgd2017} showed that the BIM can be significantly improved by starting
from a random point within the $\epsilon$ norm ball.
This attack is often called \textbf{projected gradient descent}, but this name
is somewhat confusing because (1) the term ``projected gradient descent'' already
refers to an optimization method more general than the specific use for adversarial
attack, (2) the other attacks use the gradient and perform project in the same
way (the attack is the same as BIM except for the starting point) so the name
doesn't differentiate this attack from the others.

\runinhead{Carlini and Wagner attack (C\&W).}
N. Carlini and D. Wagner followed a path of L-BFGS attack.
They designed a loss function which has smaller values on adversarial examples and higher on clean examples
and searched for adversarial examples by minimizing it~\cite{CarliniWagnerAttack}.
But unlike~\cite{Szegedy-ICLR2014} they used Adam~\cite{kingma2014adam} to solve the optimization problem
and dealt with box constraints either by change of variables (i.e. $x = 0.5(\tanh(w) + 1)$)
or by projecting results onto box constraints after each step.

They explored several possible loss functions and achieved the strongest $L_2$ attack 
with following:

\begin{equation}
\|x^{adv} - x\|_p + c \max\bigl(\max_{i \neq Y}f(x^{adv})_{i} - f(x^{adv})_{Y}, -\kappa \bigr)  \rightarrow \text{minimum}
\end{equation}
where $x^{adv}$ parametrized $0.5(\tanh(w) + 1)$; $Y$ is a shorter notation for target class $y_{target}$; $c$ and $\kappa$ are method parameters.

\runinhead{Adversarial transformation networks (ATN).}
Another approach which was explored in~\cite{ATN2017} is to train a generative model to craft adversarial examples.
This model takes a clean image as input and generates a corresponding adversarial image.
One advantage of this approach is that, if the generative model itself is designed to be small,
the ATN can generate adversarial examples faster than an explicit optimization algorithm.
In theory, this approach can be faster than even the FGSM, if the ATN is designed to use
less computation is needed for running back-propagation on the target model.
(The ATN does of course require extra time to train, but once this cost has been paid
an unlimited number of examples may be generated at low cost)

\runinhead{Attacks on non differentiable systems.}
All attacks mentioned about need to compute gradients of the model under attack in order to craft adversarial examples.
However this may not be always possible, for example if model contains non-differentiable operations.
In such cases, the adversary can train a substitute model and utilize transferability of adversarial examples to perform an attack on non-differentiable system,
similar to black box attacks, which are described below.

\subsubsection{Black box attacks}

It was observed that adversarial examples generalize between different models~\cite{szegedy2104intriguing}.
In other words, a significant fraction of adversarial examples which fool one model are able to fool
a different model.
This property is called ``transferability'' and is used to craft adversarial examples in the black box
scenario.
The actual number of transferable adversarial examples could vary from a few percent to almost $100\%$
depending on the source model, target model, dataset and other factors.
Attackers in the black box scenario can train their own model on the same dataset as the target model,
or even train their model on another dataset drawn from the same distribution.
Adversarial examples for the adversary's model then have a good chance of fooling an unknown target model.

It is also possible to intentionally design models to systematically cause
high transfer rates, rather than relying on luck to achieve transfer.

If the attacker is not in the complete black box scenario but is allowed to use
probes, the probes may be used to train the attacker's own copy of the target
model~\cite{Papernot-2016-IntroTransferability,Papernot-2016-TransferabilityStudy}
called a ``substitute.''
This approach is powerful because the input examples sent as probes do not need
to be actual training examples; instead they can be input points chosen by the
attacker to find out exactly where the target model's decision boundary lies.
The attacker's model is thus trained not just to be a good classifier but to
actually reverse engineer the details of the target model, so the two models
are systematically driven to have a high amount of transfer.

In the complete black box scenario where the attacker cannot send probes,
one strategy to increase the rate of transfer is to use an ensemble of several
models as the source model for the adversarial examples ~\cite{liu2017delving}.
The basic idea is that if an adversarial example fools every model in the ensemble,
it is more likely to generalize and fool additional models.

Finally, in the black box scenario with probes, it is possible to just run
optimization algorithms that do not use the gradient 
to directly attack the target model
~\cite{Brendel2017-DecisionBasedBlackBox,Zoo2017-ZerothOrderBlackBox}.
The time required to generate a single adversarial example is generally much
higher than when using a substitute, but if only a small number of adversarial
examples are required, these methods may have an advantage because they do not
have the high initial fixed cost of training the substitute.

\iffalse Ian thinks we don't actually need this subsubsection

\subsubsection{Physical world attacks}

\begin{figure}[h]
\centering
\includegraphics[width=4in]{figures/PhysicalSystem}
\caption{Modelling physical world adversarial attack.
Physical system $g(\bullet)$ could be modelled as a combination of a sensor and digital classifier $f(\bullet)$,
in such case entire system could be attacked using know techniques
to construct adversarial examples for white box and black box systems.}
\label{fig:adv_comp:physical_system}
\end{figure}

Conceptually physical world attack in not much different from already discussed attack scenarios.
Physical world attack could be modelled as an attack on digital classifier
which is extended by some known or unknown sensor, see Figure~\ref{fig:adv_comp:physical_system}.

If digital classifier is known and sensor could be easily modelled then this becomes a form of a white box attack
on a classifier~\cite{Athalye2017-TurtleRifle,Sharif16AdvML}.

On the other hand if either classifier is unknown or it's hard to model the sensor, then such attack could be
considered a form of black box attack~\cite{PhysicalAdversarialExamples,Sharif16AdvML}.

\fi

\subsection{Overview of defenses}

No method of defending against adversarial examples is yet completely
satisfactory. This remains a rapidly evolving research area.
We given an overview of the (not yet fully succesful defense methods)
proposed so far.

Since adversarial perturbations generated by many methods look like
high-frequency noise to a human observer\footnote{This
may be because the human
perceptual system finds the high-frequency components to be more salient;
when blurred with a low pass filter, adversarial perturbations are often
found to have significant low-frequency components
}
multiple authors have suggested to use image preprocessing and denoising
as a potential defence against adversarial examples.
There is a large variation in the proposed preprocessing techniques,
like doing JPEG compression~\cite{das2017JpegDefense}
or applying median filtering and reducing precision of input data~\cite{Weilin2017-FeatureSqueezing}.
While such defences may work well against certain attacks,
defenses in this category have been shown to fail in the white box case,
where the attacker is aware of the
defense~\cite{Warren2017-BreakingEnsembleWeakDefenses}. % paper describes how to break feature squeezing
In the black box case, this defense can be effective in practice, as
demonstrated by the winning team of the defense competition.
Their defense, described
in \secref{sec:adv_comp:def1}, is an example of this family of denoising strategies.

Many defenses, intentionally or unintentionally, fall into a category called
``gradient masking.''
Most white box attacks operate by computing gradients of the model and thus
fail if it is impossible to compute useful gradients.
Gradient masking consists of making the gradient useless, either by changing
the model in some way that makes it non-differentiable or makes it have
zero gradients in most places, or make the gradients point away from the
decision boundary.
Essentially, gradient masking means breaking the optimizer without actually
moving the class decision boundaries substantially.
Because the class decision boundaries are more or less the same, defenses
based on gradient masking are highly vulnerable to black box transfer
~\cite{Papernot-2016-IntroTransferability}.
Some defense strategies (like replacing smooth sigmoid units with hard
threshold units) are intentionally designed to perform gradient masking.
Other defenses, like many forms of adversarial training, are not designed
with gradient masking as a goal, but seem to often learn to do gradient
masking when applied in practice.

Many defenses are based on detecting adversarial examples and refusing to
classify the input if there are signs of tampering~\cite{metzen2017detecting}.
This approach works long as the attacker is unaware of the detector or the attack is not strong enough.
Otherwise the attacker can construct an attack which simultaneously fools
the detector into thinking an adversarial input is a legitimate input and
fools the classifier into making the wrong classification~\cite{Carlini2017-Breaking10Detectors}.

%% working defenses

Some defenses work but do so at the cost of seriously reducing accuracy on clean
examples. For example, shallow RBF networks are highly robust to adversarial
examples on small datasets like MNIST \cite{goodfellow2014explaining} but have much worse
accuracy on clean MNIST than deep neural networks.
Deep RBF networks might be both robust to adversarial examples and accurate
on clean data, but to our knowledge no one has successfully trained one.

Capsule networks have shown robustness to white box attacks on the SmallNORB
dataset, but have not yet been evaluated on other datasets more commonly
used in the adversarial example literature \cite{hinton2018matrix}.

The most popular defense in current research papers is probably adversarial
training~\cite{szegedy2104intriguing,Goodfellow-2015-adversarial,LearningWithStrongAdversary}.
The idea is to inject adversarial examples into training process and train the model either on adversarial examples
or on mix of clean and adversarial examples.
The approach was successfully applied to large datasets~\cite{Kurakin-AdversarialMlAtScale},
and can be made more effective by using discrete vector code representations rather
than real number representations of the input \cite{thermometer_enconding2018}.
One key drawback of adversarial training is that it tends to overfit to the specific
attack used at training time.
This has been overcome, at least on small datasets, by adding noise prior to
starting the optimizer for the attack \cite{MadryPgd2017}.
Another key drawback of adversarial training is that it tends to inadvertently
learn to do gradient masking rather than to actually move the decision boundary.
This can be largely overcome by training on adversarial examples drawn from an
ensemble of several models~\cite{Tramer2017-EAT}.
A remaining key drawback of adversarial training is that it tends to overfit to
specific constraint region used to generate the adversarial examples (models
trained to resist adversarial examples in a max-norm ball may not resist
adversarial examples based on large modifications to background pixels \cite{adversarial_sphere2018}
even if the new adversarial examples
do not appear particularly challenging to a human observer).

\section{Adversarial competition}
\label{sec:adv_comp:competition}

The phenomenon of adversarial examples creates a new set of problems in machine learning.
Studying these problems is often difficult, because when a researcher proposes a new
attack, it is hard to tell whether their attack is strong, or whether they have not
implemented their defense method used for benchmarking well enough. Similarly, it is
hard to tell whether a new defense method works well or whether it has just not been
tested against the right attack.

To accelerate research in adversarial machine learning and pit many proposed attacks
and defenses against each other in order to obtain the most vigorous evaluation
possible of these methods, we decided to organize a competition.

In this competition participants are invited to submit methods which craft adversarial examples (attacks)
as well as classifiers which are robust to adversarial eaxmples (defenses).
When evaluating competition, we run all attack methods on our dataset to produce adversarial examples
and then run all defenses on all generated adversarial examples.
Attacks are ranked by number of times there were able to fool defenses
and defenses are scored by number of correctly classified examples.

\subsection{Dataset}

When making a dataset for these competition we had following requirements:

\begin{enumerate}
\item Large enough dataset and non-trivial problem, so the competition would be interesting.
\item Well known problem, so people potentially can reuse existing classifiers. (This ensures that competitors are able to focus on the adversarial
  nature of the challenge, rather than spending all their time
  coming up with a solution to the underlying task)
\item Data samples which were never used before, so participants unlikely to overfit to well known dataset.
\end{enumerate}

These requirements were satisfied by choosing image classification problem and
creating a dataset with ImageNet-compatible images~\cite{deng2009imagenet}.
To do this we collected a set of images which were never used in publicly available datasets, labelled them
using pretrained ImageNet classifier and then manually verified that these labels are correct.

The original (non-adversarial) ImageNet challenge~\cite{ImageNetChallenge2015}
is a complex and interesting problem,
and thus satisfies requirement number 1.
Additionally there are plenty of existing classifiers for ImageNet data, which satisfies requirement number 2.
Because we collected new images and labelled them instead of using the
already available dataset, our approach satisfies requirement number 3.

Overall we collected two datasets for this competitions:
\begin{itemize}
\item \textbf{DEV} dataset was released to participants in the beginning of the competition,
  so they can use it for development of their solutions. This dataset contained $1000$ images.
\item \textbf{FINAL} dataset was kept secret and was used to evaluate final submissions of participants.
  It contained $5000$ images.
\end{itemize}

\subsection{Tasks and competition rules}

Our competition had three tracks.
Each track had a different task:

\begin{itemize}
\item \textbf{Non-targeted adversarial attack.} In this track participants were invited
to submit a method which performs a
non-targeted black box attack, i.e. given an input image,
generate an adversarial image which is likely be misclassified by unknown classifier.
\item \textbf{Targeted adversarial attack.} In this track participants were invited
to submit a method which performs
a targeted black box attack, i.e. given an input image and a target class,
generate an adversarial image which is likely be misclassified as a given target class
by unknown classifier.
\item \textbf{Defense against adversarial attacks.} In this track, participants were invited
to submit a classifier which is robust to adversarial examples.
\end{itemize}

In all tracks participants submitted a program which performs the given task and
we ran this code using our evaluation infrastructure.

The submitted program was given a set of images
(and target classes in case of targeted attack) as an input,
and had to produce either an adversarial image (for attack submission)
or classification label (for defense submission) for each input image.

Input images were fed to each submission in batches of 100 images.
To limit the budget required for computation,
the rules of the competitions specified that each submission
had to process a batch of 100 images
in no more than 500 seconds.
If a submission was not done after 500 seconds, then processing of
the current batch was interrupted
and the submission was fed the next batch. Only successfully processed images were included in the score
of the submission (see details below in evaluation sub-section).

\subsection{Evaluation metrics}
\label{sec:adv_comp:evaluation_metric}

An attack got $1$ point each time it was able to fool a defense on single image.
If attack was unable to fool a defense or if it failed to generate an
adversarial version of an image within the time limit
then it got $0$ points for that image.
A defense got $1$ point for each correctly classified image and $0$ points for incorrect classification
or failure to produce a classification label.
A defense was used to score attacks only if it successfully
processed all input images, so that attackers were neither
rewarded nor penalized for making images that caused the
defender to crash or time out---the competition is purely about
classification accuracy.
Likewise, an attack was used to score defenses only if it succesfully
processed all inputs.
Points for each submission were added together and then normalized (using the same normalization
constant for all submissions), such that final scores of all submissions are in the range $[0, 1]$,
where $1$ means success on all images and $0$ means failure on all images.

Specifically, the score of each non-targeted attack $a(Image)$ was computed using
the following formula:

\begin{equation}\label{eq:adv_comp:score_nontarget}
\frac{1}{|D|N} \sum_{d \in D} \sum_{k=1}^{N} [d(a(I_k)) \ne TrueLabel_k] 
\end{equation}

The score of each targeted attack $a(Image, TargetLabel)$ was computed as:

\begin{equation}\label{eq:adv_comp:score_target}
\frac{1}{|D|N} \sum_{d \in D} \sum_{k=1}^{N} [d(a(I_k, TargetLabel_k)) = TargetLabel_k]
\end{equation}

The score of each defense $d(Image)$ was computed in the following way:

\begin{equation}\label{eq:adv_comp:score_defense}
\frac{1}{|A|N} \sum_{a \in A} \sum_{k=1}^{N} [d(a(I_k)) = TrueLabel_k] 
\end{equation}

Where:
\begin{itemize}
  \item  $A$ is the set of all attacks (targeted and non-targeted) which were used for evaluation of defenses (attacks that crashed on some inputs or ran
    out of time for some inputs were not used to evaluate defense);
\item $D$ is the set of all defenses which were used for evaluation of attacks
  (defenses that crashed on some inputs or ran out of time for some inputs
    were not used to evaluate attacks);
\item $N$ is the number of images in the dataset;
\item $I_{k}$ is the $k$-th image from the dataset; 
\item $TrueLabel{k}$ is the true label of image $I_k$;
\item $TargetLabel_{k}$ is the chosen target label for image $k$;
\item $[P]$ is an indicator function which equals to $1$ when $P$ is true,
  and $0$ when $P$ is false or undefined.
\item $d( \bullet )$ is a defending classifier. If the binary
  fails to complete execution within the time limit, the output of
    $d( \bullet )$ is a null label that never equals the true label.
    If $d(\bullet)$ is called on an undefined image, it is defined
    to always return the true label, so an attacker that crashes
    receives zero points.
\end{itemize}

Additionally to metrics used for ranking, after the competition we computed worst case score for each submission in defense and non-targeted attack tracks.
These scores were useful to understand how submissions act in the worst case.
To compute worst score of defense we computed accuracy of the defense against each attack and chosen minimum:

\begin{equation}
\frac{1}{N} \min_{a \in A} \sum_{k=1}^{N} [d(a(I_k)) = TrueLabel_k] 
\end{equation}

To compute worst case score of non-targeted attack we computed how often attack caused misclassification when used against
each defense and chosen minimum misclassification rate:

\begin{equation}
\frac{1}{N} \min_{d \in D} \sum_{k=1}^{N} [d(a(I_k)) \ne TrueLabel_k] 
\end{equation}

Worst case score of targeted attack could be computed in a similar way,
but generally not useful because targeted attacks are much weaker than non-targeted
and all worst scores of targeted attacks were $0$.

\subsection{Competition schedule}

The competition was announced in May 2017, launched in the beginning of July 2017 and finished on October 1st, 2017.
The ompetition was run in multiple rounds. There were three development rounds followed by the final round:
\begin{itemize}
\item August 1, 2017 - first development round
\item September 1, 2017 - second development round
\item September 15, 2017 - third development round
\item October 1, 2017 - deadline for final submission
\end{itemize}

Development rounds were optional and their main purpose was to help participants to
test their solution.
Only the final round was used to compute final scores of submissions and determine winners.

All rounds were evaluated in a similar way.
For the evaluation of the round we gathered all submissions which were submitted
before the round deadline,
ran all of them and computed scores as described in section~\ref{sec:adv_comp:evaluation_metric}.

We used DEV dataset to compute scores in development rounds
and secret FINAL dataset to compute scores in the final round.

\subsection{Technical aspects of evaluation}

\begin{algorithm}[t]
\caption{Algorithm of work of evaluation infrastructure}\label{alg:adv-comp:eval}
\begin{algorithmic}[1]
\Statex \(\triangleright\) PREPARE DATASET
\parState{Split dataset $D = \{I_1, \ldots, I_N\}$ into batches $\{B_1, \ldots, B_k\}$,
  such that each batch $B_i$ contains $100$ image $\{I_{100(i-1)+1}, \ldots, I_{100i} \}$.}
\parState{Assign size of maximum allowed perturbation $\epsilon_i$ to each batch $B_i$.
  Value of $\epsilon_i$ is randomly chosen from the set
  $\{\frac{4}{255}, \frac{8}{255}, \frac{12}{255}, \frac{16}{255}\}$}
\Statex \(\triangleright\) RUN ALL ATTACKS
\ForAll{$b \in \{1, \ldots, k\}$} \Comment{loop over all batches, $b$ is batch index}
  \ForAll{non-targeted attacks $a$}
    \parState{Run attack $a$ on batch $B_b$ and generate a batch of adversarial images $\hat{B}^{a}_{b}$.
      Size of maximum perturbation $\epsilon_b$ is provided to an attack.}
    \parState{Project each adversarial image from $\hat{B}^{a}_{b}$
      into $L_{\infty}$ $\epsilon_b$-neighborhood of corresponding clean image from $B_{b}$.}
  \EndFor
  \ForAll{targeted attacks $t$}
    \parState{Run attack $t$ on batch $B_b$ and generate a batch of adversarial images $\hat{B}^{t}_{b}$.
      Attack is provided with size of maximum perturbation $\epsilon_b$ as well as target classes for
      each image from the batch $B_b$.}
    \parState{Project each adversarial image from $\hat{B}^{t}_{b}$
      into $L_{\infty}$ $\epsilon_b$-neighborhood of corresponding clean image from $B_{b}$.}
  \EndFor
\EndFor
\Statex \(\triangleright\) RUN ALL DEFENSES
\ForAll{$b \in \{1, \ldots, k\}$} \Comment{loop over all batches, $b$ is batch index}
  \ForAll{defense $d$}
    \ForAll{non-targeted attacks $a$}
      \parState{Run defense $d$ on all images from batch $\hat{B}^{a}_{b}$}
    \EndFor
    \ForAll{targeted attacks $t$}
      \parState{Run defense $d$ on all images from batch $\hat{B}^{t}_{b}$}
    \EndFor
  \EndFor
\EndFor
\Statex \(\triangleright\) COMPUTE SCORES
\parState{Determine subset of targeted and non-targeted attacks $A$ which produces all adversarial images}
\parState{Determine subset of defenses $D$ which output classification labels for all input images}
\parState{Compute scores of all submissions using
  equations~\ref{eq:adv_comp:score_nontarget},~\ref{eq:adv_comp:score_target},~\ref{eq:adv_comp:score_defense}}
\end{algorithmic}
\end{algorithm}

Competition participants were submitting pieces of code and we have run them ourselves.
This approach posess several challanges. First of all we need to protect competition infrastructure
from malicious code. Secondly, given the dataset size and number of submissions we had to run it
in an efficient way.

We partnered with Kaggle\footnote{\url{www.kaggle.com}} and used their platform as a frontend for the competition.
Kaggle hosted competition web-site, leaderboard
and participants were uploading submissions through Kaggle.
For evaluation of each round we took all submissions from Kaggle and fed them into our evaluation infrastructure.

Evaluation infrastructure worked as described in algorithm~\ref{alg:adv-comp:eval}.
As could be seen from the algorithm, attacks could be run independently of each other.
Same for defenses. 
We took advantage of this fact and parallelized execution of all attacks and all defenses
by spreading them across multiple machines.

For final evaluation we used 100 Google Cloud VMs. At any given moment
one VM was running either one attack on one batch from the dataset
or one defense on one batch of adversarial images.
Submissions were run inside Docker containers to isolate submissions from our evaluation
infrastructure and from the outside world. 

\section{Competition results}
\label{sec:adv_comp:submissions}

\begin{figure}[t]
  \centering
  \begin{subfigure}[b]{0.45\textwidth}
    \includegraphics[width=\textwidth]{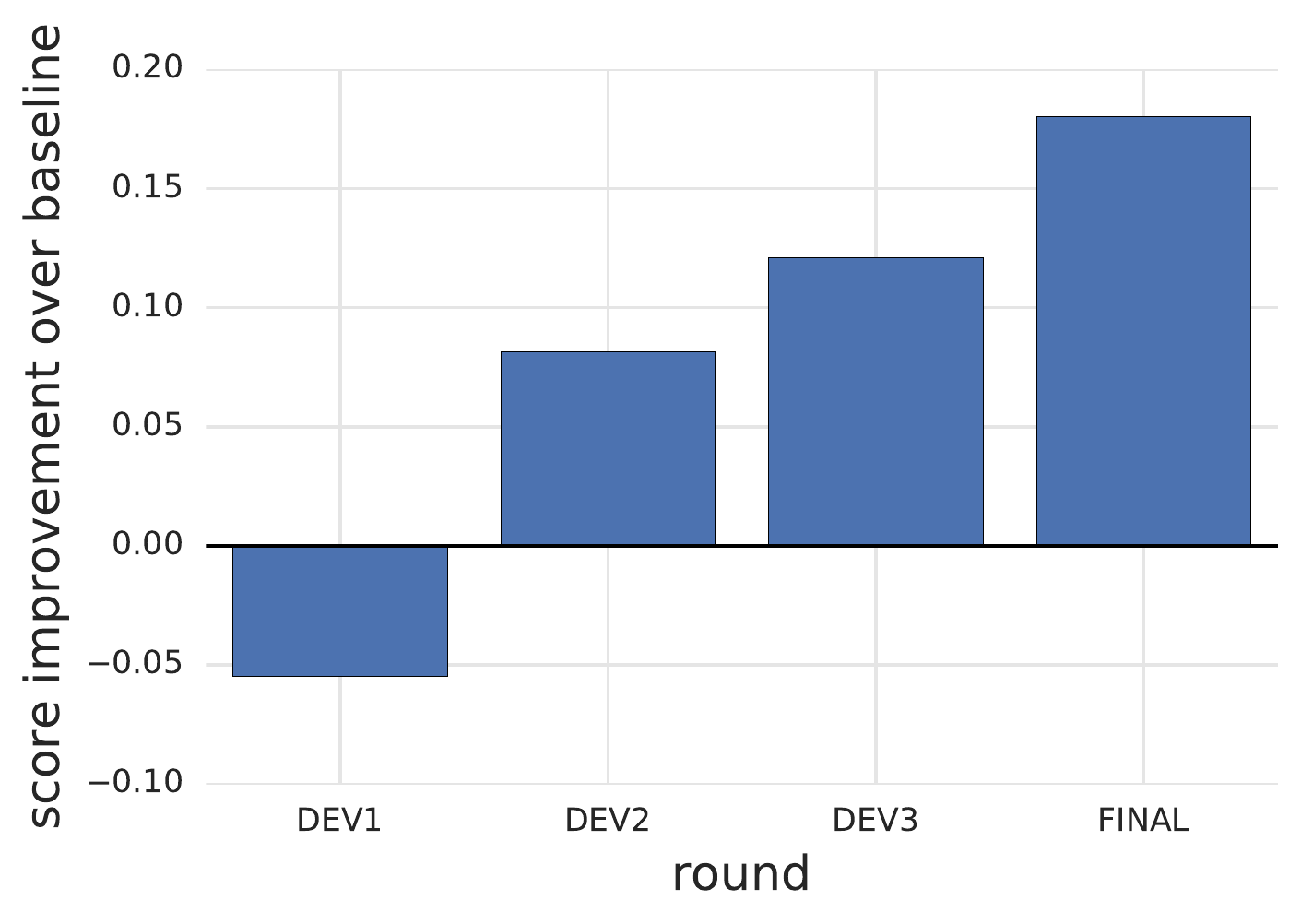}
    \caption{Defenses}
  \end{subfigure}
  \,
  \begin{subfigure}[b]{0.45\textwidth}
    \includegraphics[width=\textwidth]{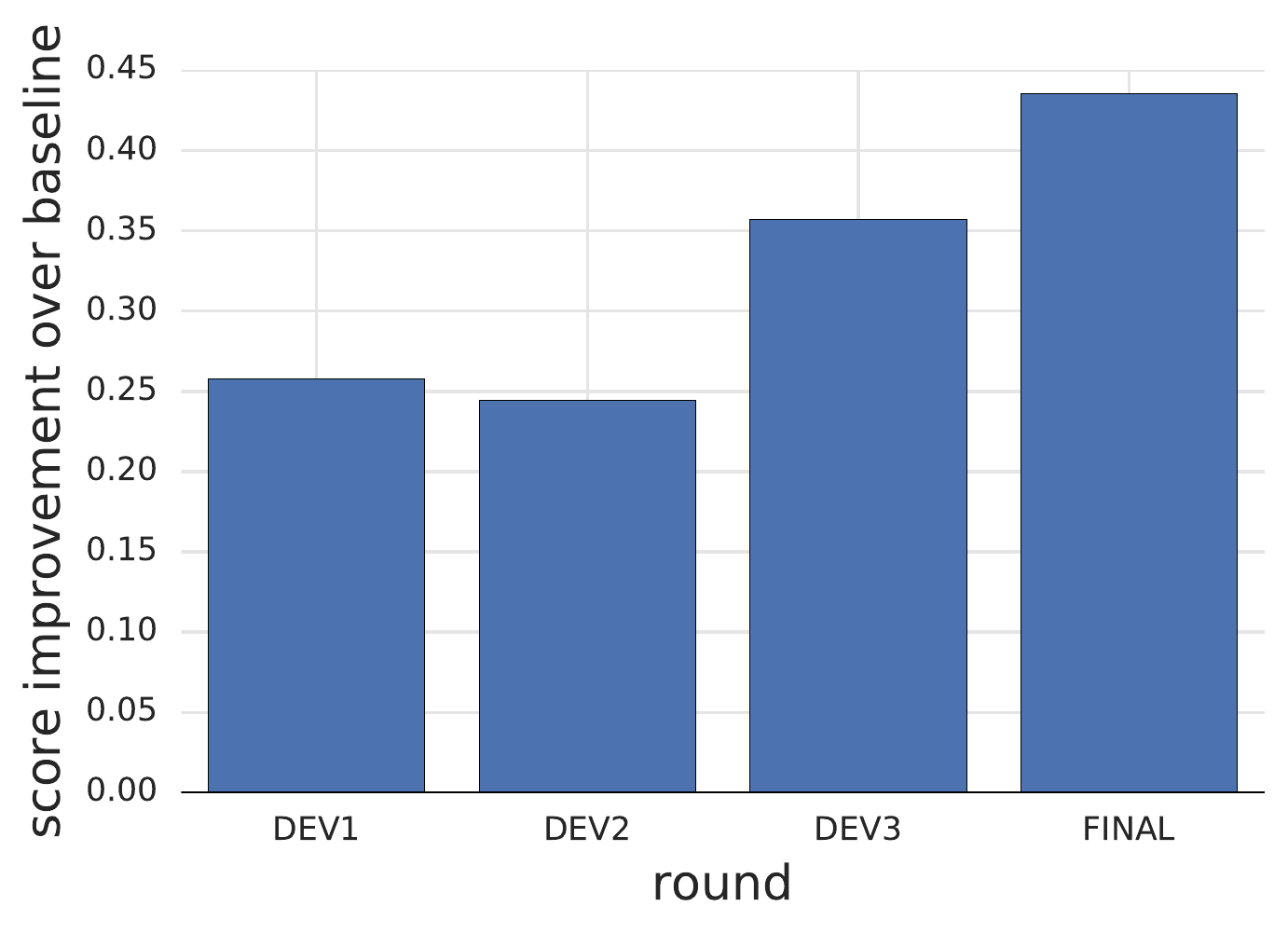}
    \caption{Non-targeted attacks}
  \end{subfigure}
  \,
  \begin{subfigure}[b]{0.45\textwidth}
    \includegraphics[width=\textwidth]{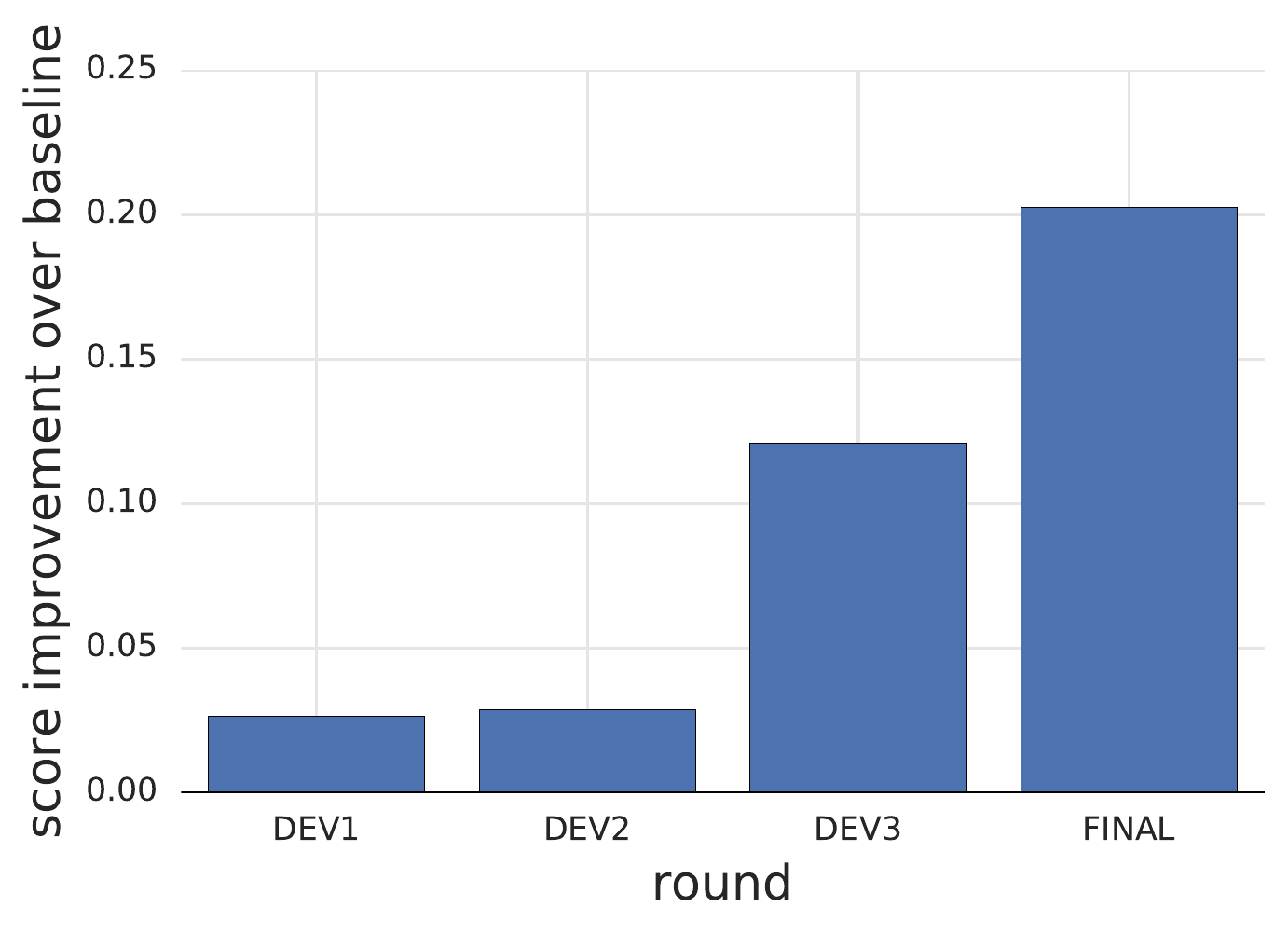}
    \caption{Targeted attacks}
  \end{subfigure}
  \caption{Plots which shows difference between score of top submission and best baseline in each round in each track.
  As could be seen from the plot, submissions kept improving each round.
  }\label{fig:adv_comp:improvements}
\end{figure}

For the final round we had 91 non-targeted attack submissions,
65 targeted attack submission and 107 defense submissions.
During the course of competitions submissions kept improving their results compared to baselines which could be
seen from Figure~\ref{fig:adv_comp:improvements}.

\begin{figure}[t]
  \centering
  \begin{subfigure}[b]{0.45\textwidth}
    \includegraphics[width=\textwidth]{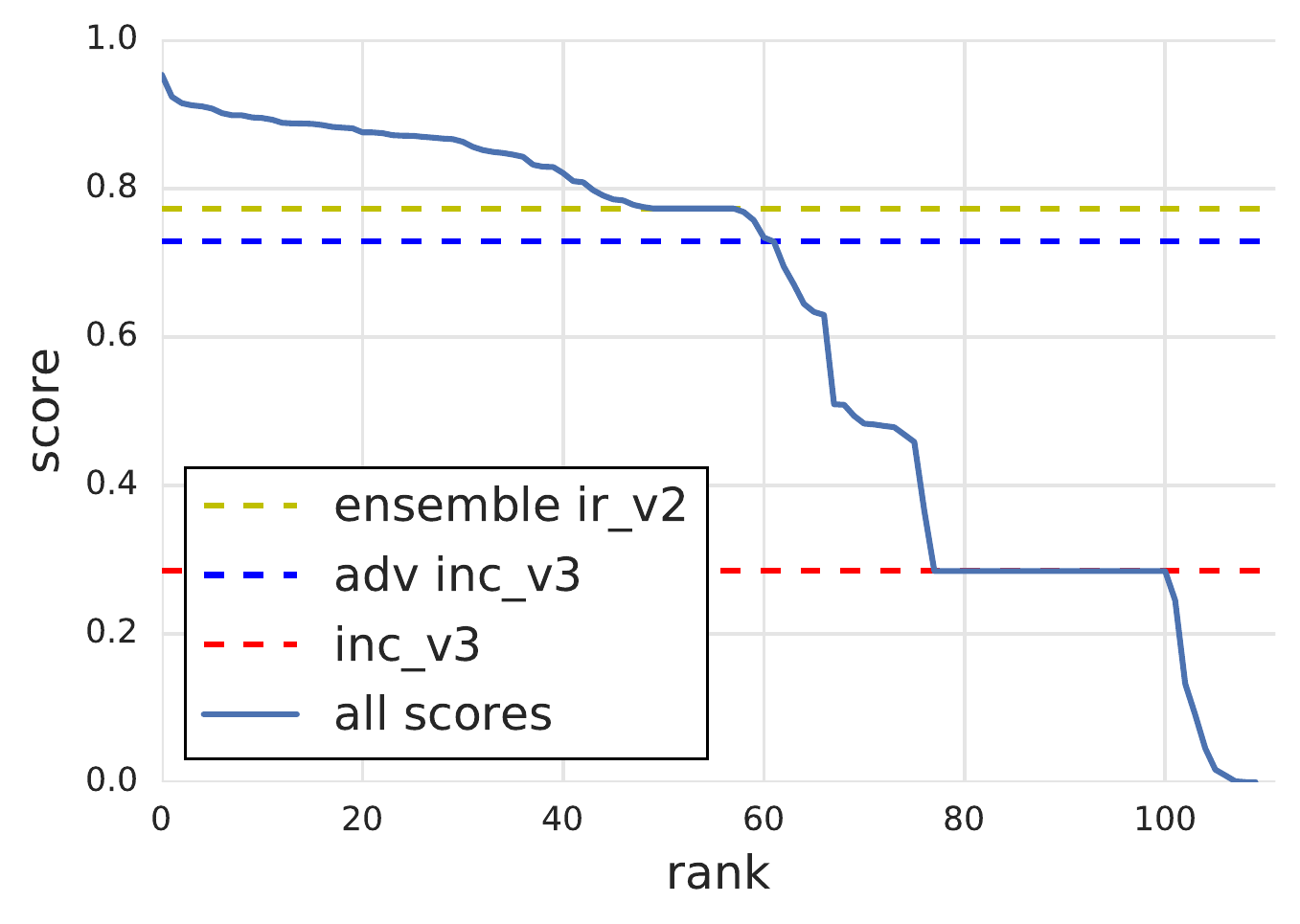}
    \caption{Defenses}
  \end{subfigure}
  \,
  \begin{subfigure}[b]{0.45\textwidth}
    \includegraphics[width=\textwidth]{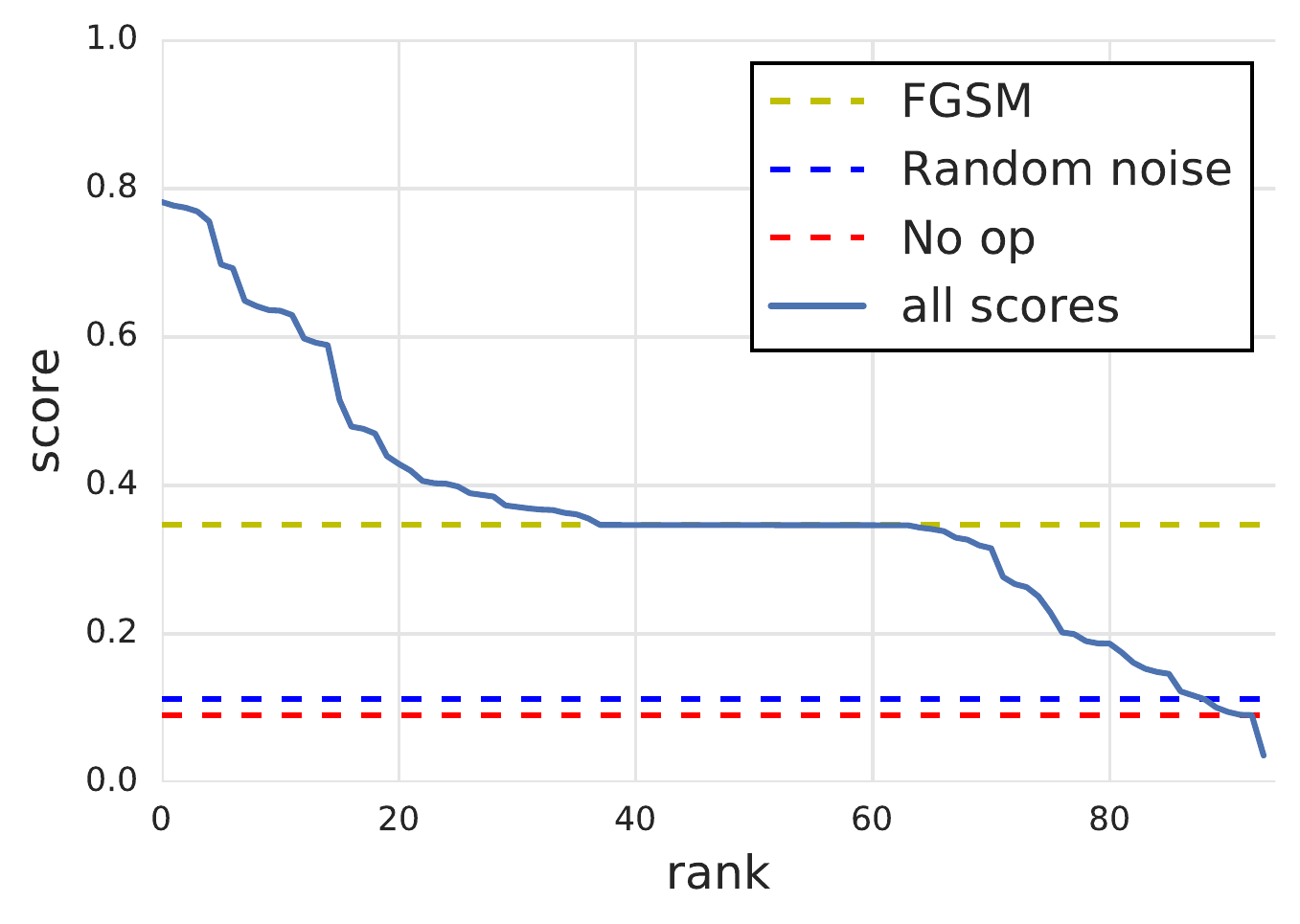}
    \caption{Non-targeted attacks}
  \end{subfigure}
  \,
  \begin{subfigure}[b]{0.45\textwidth}
    \includegraphics[width=\textwidth]{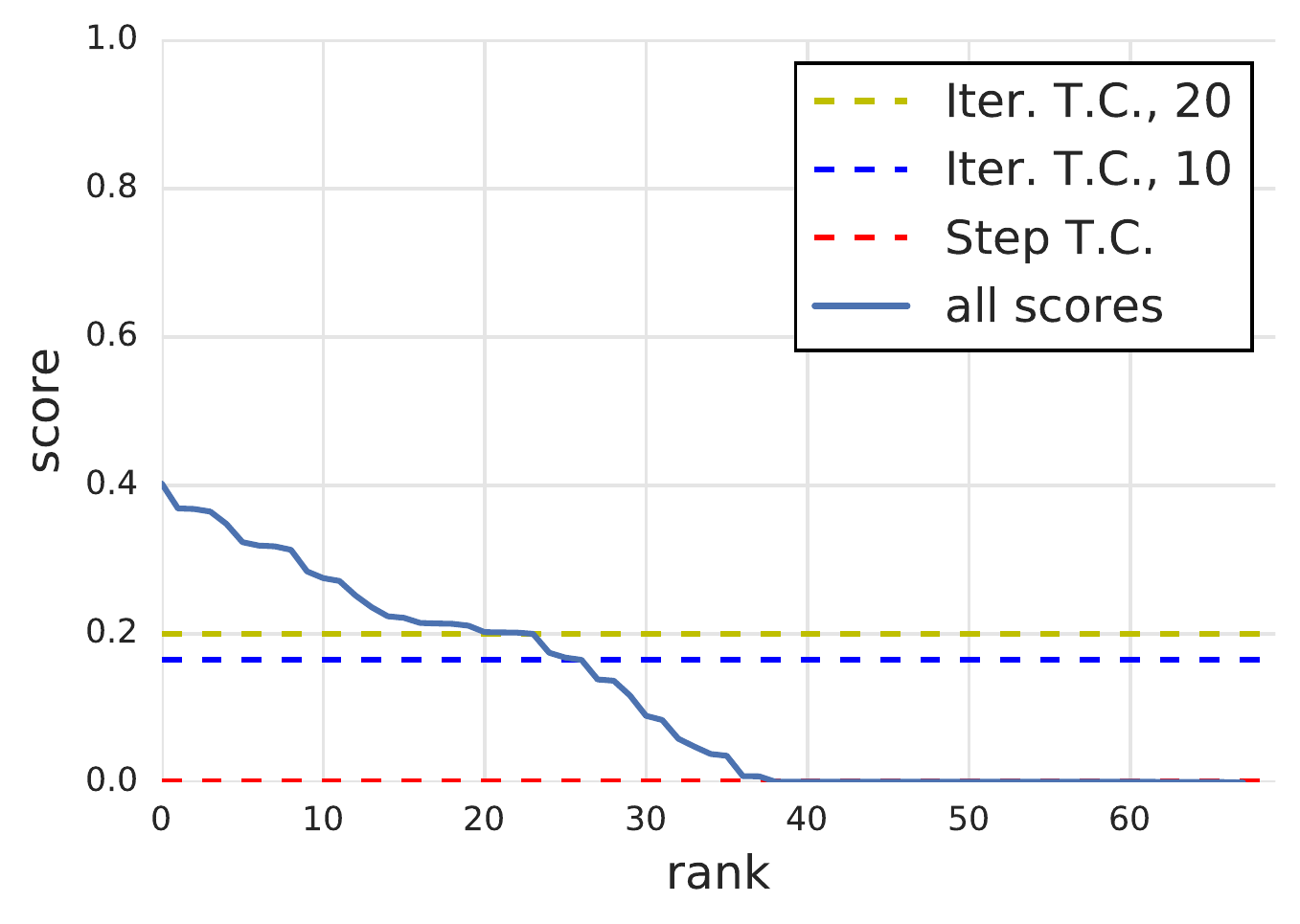}
    \caption{Targeted attacks}
  \end{subfigure}
  \caption{Plots with scores of submissions in all three tracks.
  Solid line of each plot is scores of submissions depending on submission rank.
  Dashed lines are scores of baselines we provided.
  These plots demonstrate difference between best and worst submissions as well as
  how much top submissions were able to improve provided baselines.
  }\label{fig:adv_comp:scores_plot}
\end{figure}

Final results of the top submissions in each track are provided in tables~\ref{tab:adv_comp:defenses},
\ref{tab:adv_comp:nontarget} and~\ref{tab:adv_comp:target}.
Meaning of the columns is following.
\textbf{Rank} is submission rank in final scoring,
\textbf{score} is submission score as described in Section~\ref{sec:adv_comp:evaluation_metric},
\textbf{raw score} is un-normalized score which is number of times submission got a point on the image,
\textbf{worst score} is submission score in the worst case and \textbf{medial eval time} is 
median time needed for evaluation of one batch of $100$ images.
To put things into prospective, plots of all submission scores in final round from best to worst and comparison with
provided baselines are depicted in Figure~\ref{fig:adv_comp:scores_plot}.

\begin{table}
\centering
\caption{Top-5 defense submissions, best baseline and submission with maximum worst-case score}
\label{tab:adv_comp:defenses}
\begin{tabular}{r|l|r|r|r|r}
\hline\noalign{\smallskip}
 Rank &                 Team name or baseline &  Score &    Raw Score & Worst Score &  Median eval time \\
\noalign{\smallskip}\svhline\noalign{\smallskip}
     1 &                          TSAIL &         0.953164 &  691044 &      0.1184 &            51.0 \\
     2 &                         iyswim &         0.923524 &  669555 &      0.2520 &           126.0 \\
     3 &                    Anil Thomas &         0.914840 &  663259 &      0.2666 &            97.0 \\
     4 &                           erko &         0.911961 &  661172 &      0.2920 &            87.0 \\
     5 &               Stanford \& Suns &         0.910593 &  660180 &      0.0682 &           129.0 \\
    24 &                  Rafael Moraes &         0.871739 &  632011 &      \textbf{0.5358} &            17.0 \\
    56 &  Baseline (Ens. adv. ir\_v2) &         0.772908 &  560358 &      0.0186 &            17.0 \\
\noalign{\smallskip}\hline\noalign{\smallskip}
\end{tabular}
\end{table}

\begin{table}
\centering
\caption{Top-5 non-targeted attack submissions, best baseline and best submission with according to worst-case score.}
\label{tab:adv_comp:nontarget}
\begin{tabular}{r|l|r|r|r|r}
\hline\noalign{\smallskip}
 Rank & Team name or baseline &  Score &  Raw Score &  Worst Score &  Median eval time \\
\noalign{\smallskip}\hline\noalign{\smallskip}
     1 &         TSAIL   &         0.781644 &  410363 &      0.1364 &           423.0 \\
     2 &         Sangxia &         0.776855 &  407849 &      \textbf{0.3412} &           421.0 \\
     3 & Stanford \& Sun &         0.774025 &  406363 &      0.2722 &           497.0 \\
     4 &         iwiwi   &         0.768981 &  403715 &      0.1352 &            76.0 \\
     5 &        toshi\_k &         0.755598 &  396689 &      0.3322 &           448.0 \\
    44 &  Baseline (FGSM)  &         0.346400 &  181860 &      0.0174 &            17.0 \\
\noalign{\smallskip}\hline\noalign{\smallskip}
\end{tabular}
\end{table}

\begin{table}
\centering
\caption{Top-5 targeted attack submissions and best baseline.}
\label{tab:adv_comp:target}
\begin{tabular}{r|l|r|r|r}
\hline\noalign{\smallskip}
 Rank &                Team &  Score &     Raw Score &  Median Eval Time \\
\noalign{\smallskip}\hline\noalign{\smallskip}
     1 &           TSAIL &         0.402211 &  211161.0 &           392.0 \\
     2 &         Sangxia &         0.368773 &  193606.0 &           414.0 \\
     3 &      FatFingers &         0.368029 &  193215.0 &           493.0 \\
     4 &     Anil Thomas &         0.364552 &  191390.0 &           495.0 \\
     5 &             WNP &         0.347935 &  182666.0 &           487.0 \\
    24 &  Baseline (Iter. T. C. 20) &         0.199773 &  104881.0 &           127.0 \\
\noalign{\smallskip}\hline\noalign{\smallskip}
\end{tabular}
\end{table}

As could be seen from the tables, best defenses achieved more than $90\%$ accuracy
on all adversarial images from all attacks.
At the same time worst case scores of defenses are much lower.
The highest worst case score among all defenses is only $53.6\%$.

This could be an indication that it's possible to achieve pretty high accuracy against adversarial examples in average case,
however model will still be suspectible to adversarial examples and could be fooled if adversary will be able to find them.

Similar observation holds for attacks.
Best attacks achieved up to $78\%$ fooling rate against all defenses, at the same time worst case score of attacks was no more than $34\%$.

\section{Top scoring submissions}

In the remainder of the chapter, we present the solutions of
several of the top-scoring teams.

To describe the solutions, we use the following notation:
\begin{itemize}
  \item $x$ - input image with label $y_{true}$. Different images are distinguished by superscripts,
   for examples images $x^{1}, x^{2}, \ldots$ with labels $y^{1}_{true}, y^{2}_{true}, \ldots$.
  \item $y_{target}$ is a target class for image $x$ for targeted adversarial attack.
  \item Functions with names like $f(\bullet), g(\bullet), h(\bullet), \ldots$ are classifiers
    which map input images into logits.
    In other words $f(x)$ is logits vector of networks $f$ on image $x$
  \item $J(f(x), y)$ - cross entropy loss between logits $f(x)$ and class $y$.
  \item $\epsilon$ - maximum $L_{\infty}$ norm of adversarial perturbation.
  \item $x_{adv}$ - adversarial images.
    For iterative methods $x_{adv}^{i}$ is adversarial example generated on step $i$.
  % Notation from team iwiwi:
  \item $Clip_{[a, b]}(\bullet)$ is a function which performs element-wise clipping
    of input tensor to interval $[a, b]$.
  \item $\mathcal{X}$ is the set of all training examples.
\end{itemize}

All values of images are normalized to be in $[0, 1]$ interval.
Values of $\epsilon$ are also normalized to $[0, 1]$ range,
for examples $\epsilon = \frac{16}{255}$ correspond
to uint8 value of epsilon equal to $16$.

%%%%%%%%%%%%%%%%%%%%%%%%%%%%%%%%%%%%%%%%%%%%%%%%%%%%%%%%%%%%%%%%%%%%%%%%%%%%%%%
% All submissions
%%%%%%%%%%%%%%%%%%%%%%%%%%%%%%%%%%%%%%%%%%%%%%%%%%%%%%%%%%%%%%%%%%%%%%%%%%%%%%%

%%%%%%%%%%%%%%%%%%%%%%%%%%%%%%%%%%%%%%%%%%%%%%%%%%%%%%%%%%%%%%%%%%%
%
% Subsection about submission XXXXXXXX
%
%%%%%%%%%%%%%%%%%%%%%%%%%%%%%%%%%%%%%%%%%%%%%%%%%%%%%%%%%%%%%%%%%%%

\subsection{1st place in defense track: team TsAIL}
\label{sec:adv_comp:def1}

\runinhead{Team members:} Yinpeng Dong, Fangzhou Liao, Ming Liang, Tianyu Pang, Jun Zhu and Xiaolin Hu.

In this section, we introduce the high-level representation guided denoiser~(HGD) method, which won the first place in the defense track.
The idea is to train a neural network based denoiser to remove the adversarial perturbation.

\subsubsection{Dataset} \label{sec:hgd:dataset}
To prepare the training set for the denoiser, we first extracted 20K images from the ImageNet training set (20 images per class).
Then we used a bunch of adversarial attacks to distort these images and form a training set.
Attacking methods included FGSM and I-FGSM and were applied to the many models and their ensembles to simulate weak and strong attacks.

\subsubsection{Denoising U-net}
Denoising autoencoder (DAE) \cite{vincent2008extracting} is a potential choice of the denoising network. 
But DAE has a bottleneck for the transmission of fine-scale information between the encoder and decoder.
This bottleneck structure may not be capable of carrying the multi-scale information contained in the images.
That's why we used a denoising U-net (DUNET). 

Compared with DAE, the DUNET adds some lateral connections from encoder layers to their corresponding decoder layers of the same resolution. 
In this way, the network is learning to predict adversarial noise only, which is more relevant to denoising and easier than reconstructing the whole image \cite{zhang2017beyond}.
The clean image can be readily obtained by subtracting the noise from the corrupted input:
\begin{equation}
d\hat{x} = D_w(x^{adv}).
\end{equation}
\begin{equation}
\hat{x} = x^{adv}-d\hat{x}.
\end{equation}
where $D_w$ is a denoiser network with parameters $w$, $d\hat{x}$ is predicted adversarial noise and $\hat{x}$ is reconstructured clean image.

\subsubsection{Loss function} \label{sec:hgd:method}

The vanilla denoiser uses the reconstructing distance as the loss function, but we found a better method.
Given a target neural network, we extract its representation at $l$-th layer for $x$ and $\hat{x}$, and calculate the loss function as:
\begin{equation}
L=\|f_l(\hat{x}) - f_l(x)\|_1.
\end{equation}
The corresponding models are called HGD, because the supervised signal comes from certain high-level layers of the classifier and carries guidance information related to image classification. 

We propose two HGDs with different choices of $l$.
For the first HGD, we define $l=-2$ as the index of the topmost convolutional layer.
This denoiser is called feature guided denoiser (FGD).
For the second HGD, we use the logits layer.
So it is called logits guided denoiser (LGD). 

Another kind of HGD uses the classification loss of the target model as the denoising loss function, which is supervised learning as ground truth labels are needed.
This model is called class label guided denoiser (CGD).
In this case the loss function is optimized with respect to the parameters of the denoiser $w$,
while the parameters of the guiding model are fixed. 

Please refer to our full-length paper \cite{liao2017defense} for more information.
 % 1st place defense - team TsAIL
%%%%%%%%%%%%%%%%%%%%%%%%%%%%%%%%%%%%%%%%%%%%%%%%%%%%%%%%%%%%%%%%%%%
%
% Subsection about submission XXXXXXXX
%
%%%%%%%%%%%%%%%%%%%%%%%%%%%%%%%%%%%%%%%%%%%%%%%%%%%%%%%%%%%%%%%%%%%

\subsection{1st place in both attack tracks: team TsAIL}
%\subsection{Momentum iterative attack method}
\label{sec:adv_comp:mim}

\runinhead{Team members:} Yinpeng Dong, Fangzhou Liao, Ming Liang, Tianyu Pang, Jun Zhu and Xiaolin Hu.

In this section, we introduce the momentum iterative gradient-based attack method, which won the first places in both the non-targeted attack and targeted attack tracks.
We first describe the algorithm in Sec.~\ref{sec:mim:method},
and then illustrate our submissions for non-targeted and targeted attacks respectively in Sec.~\ref{sec:mim:non-target}
and Sec.~\ref{sec:mim:target}. A more detailed description can be found in~\cite{dong2017boosting}.

\subsubsection{Method}
\label{sec:mim:method}

The momentum iterative attack method is built upon the basic iterative method~\cite{Kurakin-PhysicalAdversarialExamples}, by adding a momentum term to greatly improve the transferability of the generated adversarial examples.

Existing attack methods exhibit low efficacy when attacking black-box models, due to the well-known trade-off between the attack strength and the transferability~\cite{Kurakin-AdversarialMlAtScale}.
In particular, one-step method (e.g., FGSM) calculates the gradient only once using the assumption of linearity of the decision boundary around the data point.
However in practice, the linear assumption may not hold when the distortions are large~\cite{liu2017delving}, which makes the adversarial examples generated by one-step method  ``underfit'' the model,
limiting attack strength.
In contrast, basic iterative method greedily moves the adversarial example in the direction of the gradient in each iteration. Therefore, the adversarial example can easily drop into poor local optima and ``overfit'' the model, which are not likely to transfer across models.

In order to break such a dilemma, we integrate momentum~\cite{Poljak1964Some} into the basic iterative method for the purpose of stabilizing update directions and escaping from poor local optima, which are the common benefits of momentum in optimization literature~\cite{Korczak1998Optimization,Sutskever2013On}. As a consequence, it alleviates the trade-off between the attack strength and the transferability, demonstrating strong black-box attacks.

The momentum iterative method for non-targeted attack is summarized as:
\begin{equation}
g^{t+1} = \mu \cdot g^{t} + \frac{\nabla_{x}J(f(x^{t}_{adv}),y_{true})}{\|\nabla_{x}J(f(x^{t}_{adv}),y_{true})\|_1}, \hspace{1ex}
x^{t+1}_{adv} = \mathrm{Clip}_{[0, 1]}(x^{t}_{adv} + \alpha\cdot\mathrm{sign}(g^{t+1}))
\end{equation}
where $g^{0}=0$, $x^{0}_{adv}=x$, $\alpha = \frac{\epsilon}{T}$ with T being the number of iterations.
$g^t$ gathers the gradients of the first $t$ iterations with a decay factor $\mu$
and adversarial example $x^t_{adv}$  is perturbed in the direction of the sign of $g^t$ with the step size $\alpha$.
In each iteration, the current gradient $\nabla_{x}J(f(x^{t}_{adv}),y_{true})$
is normalized to have unit $L_1$ norm (however other norms will work too),
because we noticed that the scale of the gradients varies in magnitude between iterations.

To obtain more transferable adversarial examples, we apply the momentum iterative method to attack an ensemble of models.
If an example remains adversarial for multiple models, it may capture an intrinsic direction that always fools these models
and is more likely to transfer to other models at the same time~\cite{liu2017delving}, thus enabling powerful black-box attacks.

We propose to attack multiple models whose \textit{logit} activations are fused together, because the logits capture the logarithm relationships between the probability predictions, an ensemble of models fused by logits aggregates the fine detailed outputs of all models, whose vulnerability can be easily discovered.
Specifically, to attack an ensemble of $K$ models, we fuse the logits as 
\begin{equation}\label{eq:mim:logits_fusing}
f(x) = \sum_{k=1}^K w_k f_k(x)
\end{equation}
where $f_k(x)$ are the $k$-th model, $w_k$ is the ensemble weight with $w_k \geq 0$ and $\sum_{k=1}^K w_k = 1$. Therefore we get a big ensemble model $f(x)$ and we can use the momentum iterative method to attack $f$.

\subsubsection{Submission for non-targeted attack}
\label{sec:mim:non-target}

In non-targeted attack, we implemented the momentum iterative method for attacking an ensemble of
following models:
\begin{itemize}
\item Normally trained (i.e. without adversarial training)
  Inception~v3~\cite{Szegedy2016Rethinking}, Inception~v4~\cite{szegedy2017inception},
  Inception~Resnet~v2~\cite{szegedy2017inception} and Resnet~v2-101~\cite{he2016identity} models.
\item Adversarially trained Inception~v3\textsubscript{adv}~\cite{Kurakin-AdversarialMlAtScale} model.
\item Ensemble adversarially trained Inc-v3\textsubscript{ens3}, Inc-v3\textsubscript{ens4} and IncRes-v2\textsubscript{ens} models from~\cite{Tramer2017-EAT}.
\end{itemize}

Ensemble weights (from Equation~\ref{eq:mim:logits_fusing}) were
$\frac{0.25}{7.25}$ for Inception-v3\textsubscript{adv} and $\frac{1}{7.25}$ for all other models.
The number of iterations was $10$ and the decay factor $\mu$ was $1.0$.

\subsubsection{Submission for targeted attack}\label{sec:mim:target}

For targeted attacks, we used a different formula of momentum iterative method:
\begin{align}
g^{t+1} & = \mu \cdot g^{t} + \frac{\nabla_{x}J(f(x^{t}_{adv}),y_{target})}{\mathrm{std}(\nabla_{x}J(f(x^{t}_{adv}),y_{target})}\\
x^{t+1}_{adv} & = \mathrm{Clip}_{[0,1]}\Bigl(x^{t}_{adv} - \alpha\cdot \mathrm{Clip}_{[-2, 2]}(\mathrm{round}(g^{t+1})) \Bigr)
\end{align}
where $\mathrm{std}(\bullet)$ is the standard deviation and $\mathrm{round}(\bullet)$ is rounding to nearest integer.
Values of $\mathrm{Clip}_{[-2, 2]}(\mathrm{round}(\bullet))$ are in set $\{-2, -1, 0, 1, 2\}$
which enables larger search space compared to sign function.

No transferability of the generated adversarial examples was observed in the targeted attacks,
so we implement our method for attacking several commonly used white-box models.

We built two versions of the attacks.
If the size of perturbation $\epsilon$ was smaller than $\frac{8}{255}$,
we attacked ensemble of Inception v3 and IncRes-v2\textsubscript{ens}
with weights $\frac{1}{3}$ and $\frac{2}{3}$;
otherwise we attacked an ensemble of Inception v3, Inception-v3\textsubscript{adv},
Inc-v3\textsubscript{ens3}, Inc-v3\textsubscript{ens4} and IncRes-v2\textsubscript{ens} with ensemble weights $\frac{4}{11}, \frac{1}{11}, \frac{1}{11}, \frac{1}{11}$ and $\frac{4}{11}$.
The number of iterations were $40$ and $20$ respectively, and the decay factor $\mu$ was $1.0$. 
 % 1st place both attacks - team TsAIL
%%%%%%%%%%%%%%%%%%%%%%%%%%%%%%%%%%%%%%%%%%%%%%%%%%%%%%%%%%%%%%%%%%%
%
% Subsection about submission: mitigating adversarial effects through randomization
%
%%%%%%%%%%%%%%%%%%%%%%%%%%%%%%%%%%%%%%%%%%%%%%%%%%%%%%%%%%%%%%%%%%%

\subsection{2nd place in defense track: team iyswim}
%\subsection{Mitigating adversarial effects through randomization}
\label{sec:adv_comp:mitigating_adv}

\runinhead{Team members:} Cihang Xie, Jianyu Wang, Zhishuai Zhang, Zhou Ren and Alan Yuille

\begin{figure}[h!]
\centering
\vspace{-0.2cm}
\includegraphics[width=1.0\columnwidth]{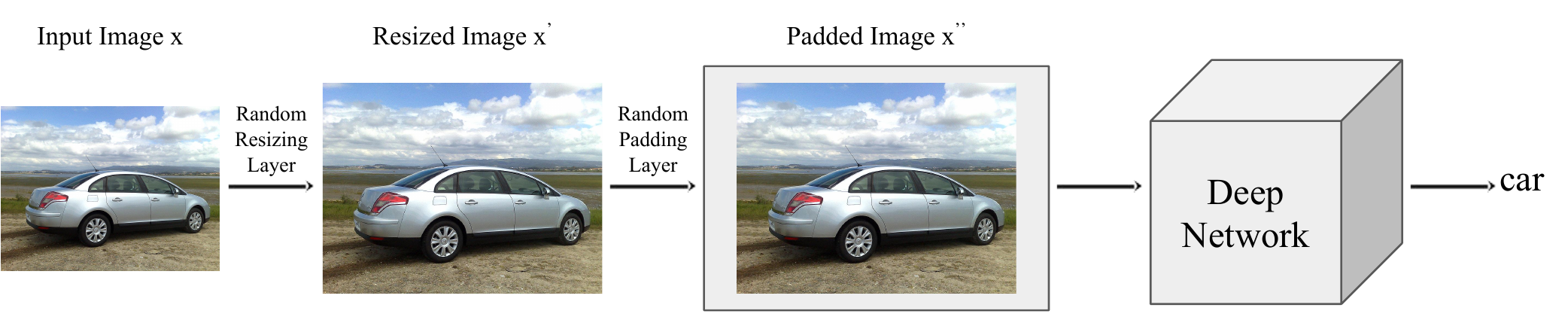}
\caption{The pipeline of the proposed defense method. The input image $x$ first goes through the random resizing layer with \textbf{a} random scale applied. Then the random padding layer pads the resized image $x^{\prime}$ in \textbf{a} random manner. The resulting padded image $x^{\prime\prime}$ is used for classification.}
\label{Fig:pipline}
\vspace{-0.5cm}
\end{figure}

In this submission, we propose to utilize randomization as a defense against adversarial examples.
Specifically, we propose a randomization-based method, as shown in figure \ref{Fig:pipline},
which adds a random resizing layer and a random padding layer to the beginning of the classification networks.
Our method enjoys the following advantages: (1) no additional training or fine-tuning; (2) very few additional computations; (3) compatible with other adversarial defense methods.
By combining the proposed randomization method with an adversarially trained model, it ranked \textbf{No.2} in the NIPS adversarial defense challenge. 

\subsubsection{Randomization as defense}
Intuitively, the adversarial perturbation generated by iterative attacks may easily get over-fitted to the specific network parameters, and thus be less transferable. Due to this weak generalization ability, we hypothesis that low-level image transformations, e.g., resizing, padding, compression, etc, may probably destroy the specific structure of adversarial perturbations, thus making it a good defense. It can even defend against white-box iterative attacks if random transformations are applied. This is because each test image goes through a transformation randomly and the attacker does not know this specific transformation when generating adversarial noise.

\subsubsection{Randomization layers}
The first randomization layer is a random resizing layer, which resizes the original input image $x$ with the size $W \times H \times 3$ to a new image $x^{\prime}$ with random size $W^\prime \times H^\prime \times 3$. Note that, $|W^\prime - W|$ and $|H^\prime - H|$ should be within a reasonably small range, otherwise the network performance on clean images would significantly drop. Taking Inception-ResNet network~\cite{szegedy2017inception} as an example, the original data input size is $299 \times 299 \times 3$. Empirically we found that the network accuracy hardly drops if we control the height and width of the resized image $x^\prime$ to be within the range $\left[299, 331\right)$.

The second randomization layer is the random padding layer, which pads  zeros around the resized image in a random manner. Specifically, by padding the resized image $x^\prime$ into a new image $x^{\prime\prime}$ with the size $W^{\prime\prime} \times H^{\prime\prime} \times 3$, we can choose to pad $w$ zero pixels on the left, $W^{\prime\prime} - W^{\prime} - w$ zero pixels on the right, $h$ zero pixels on the top and $H^{\prime\prime} - H^{\prime} - h$ zero pixels on the bottom. This results in a total number of $(W^{\prime\prime} - W^{\prime} + 1) \times (H^{\prime\prime} - H^{\prime} + 1)$ different possible padding patterns.

During implementation, the original image first goes through two randomization layers, and then we pass the transformed image to the original CNN for classification. The pipeline is illustrated in figure \ref{Fig:pipline}.

\subsubsection{Randomization layers + adversarial training}
\label{sec:random+adv_train}
Recently, adversarial training~\cite{Kurakin-AdversarialMlAtScale, Tramer2017-EAT} was developed as an effective defense for single-step attacks. Thus by adding the proposed random transformations as additional layers to an adversarially trained model~\cite{Tramer2017-EAT}, it is expected that this method is able to effectively defend against both single-step and iterative attacks, including both black-box and white-box settings. 

\subsubsection{Submission details and results}
An adversarially trained model appended with randomization layers was submitted as our defense model to the challenge. Specifically, we (1) set the resizing range to be $[310, 331)$; (2) set the padding size to be $331 \times 331 \times 3$; (3) choose the adversarially trained model to be \textit{ens-adv-Inception-Resnet-v2}\footnote{\url{https://download.tensorflow.org/models/ens_adv_inception_resnet_v2_2017_08_18.tar.gz}}; (4) average the prediction results over $30$ randomization patterns for each image; (5) flip the input image with probability $0.5$ for each randomization pattern. The whole implementation is public available\footnote{\url{https://github.com/cihangxie/NIPS2017_adv_challenge_defense}}.

By evaluating our model against $156$ different attacks, it reaches a normalized score of $0.924$ (ranked No.$2$ among $107$ defense models), which is far better than using ensemble adversarial training~\cite{Tramer2017-EAT} alone with a normalized score of $0.773$. This result further demonstrates that the proposed randomization method can effectively make deep networks much more robust to adversarial attacks.

\subsubsection{Attackers with more information}
When submitting the proposed defense method to the NIPS competition, the randomization layers are remained as an unknown network module for the attackers. We thus test the robustness of this defense method further by assuming that the attackers are aware of the existence of randomization layers. Extensive experiments are performed in \cite{xie2017mitigating}, and it shows that the attackers still cannot break this defense completely in practice. Interested readers can refer to \cite{xie2017mitigating} for more details.
 % 2nd place defense - team iyswim
%%%%%%%%%%%%%%%%%%%%%%%%%%%%%%%%%%%%%%%%%%%%%%%%%%%%%%%%%%%%%%%%%%%
%
% Subsection about submission XXXXXXXX
%
%%%%%%%%%%%%%%%%%%%%%%%%%%%%%%%%%%%%%%%%%%%%%%%%%%%%%%%%%%%%%%%%%%%

\subsection{2nd place in both attack tracks: team Sangxia}
\label{sec:adv_comp:submission_sangxia}

\runinhead{Team members:} Sangxia Huang

In this section, we present the submission by Sangxia Huang
for both non-targeted and targeted attacks. The approach is
an iterated FGSM attack against an ensemble of classifiers
with random perturbations and augmentations for increased robustness and 
transferability
of the generated attacks. The source code is available online.
\footnote{\url{https://github.com/sangxia/nips-2017-adversarial}}
We also optimize the iteration steps for improved efficiency 
as we describe in more details below.

\runinhead{Basic idea}
An intriguing property of adversarial examples 
observed in many works 
~\cite{Papernot-2016-IntroTransferability,szegedy2104intriguing,goodfellow2014explaining,Papernot-2016-TransferabilityStudy}
is that adversarial examples generated for one classifier transfer
to other classifiers.
Therefore, a natural approach for effective attacks against 
unknown classifiers is to generate
strong adversarial examples against a large collection of 
classifiers.

Let $f^{1}, \ldots, f^{k}$ be an ensemble of image classifiers
that we choose to target. In our solution we give equal weights to each of them. 
For notation simplicity, we assume that the inputs to all $f^{i}$
have the same size.  Otherwise, we first insert a bi-linear 
scaling layer, which is differentiable. The differentiability ensures that the
correct gradient signal is propagated through the scaling layer
to the individual pixels of the images.

Another idea we use to increase robustness and transferrability
of the attacks is image augmentation.
Denote by $T_{\theta}$ an image augmentation function with 
parameter $\theta$.
For instance, we can have $\theta \in [0, 2\pi)$ as an angle and 
$T_{\theta}$ as the function that rotates the input image clock-wise
by $\theta$. The parameter $\theta$ can also be a vector. For
instance, we can have $\theta \in (0,\infty)^{2}$ as scaling factors
in the width and height dimension, and $T_{\theta}$ as the function
that scales the input image in the width direction by $\theta_1$
and in the height direction by $\theta_2$. In our final algorithm,
$T_{\theta}$ takes the general form of a projective transformation
with $\theta \in \mathbb{R}^8$ as implemented in 
\verb|tf.contrib.image.transform|. 

Let $x$ be an input image, and $y_{true}$ be the label of $x$.
Our attack algorithm works to find an $x^{adv}$ that maximizes the 
expected average cross entropy loss of the predictions of
$f^{1}, \ldots, f^{k}$ over a random input augmentation
\footnote{The distribution we use for $\theta$ corresponds to a small
random augmentation. See code for details.}
\[
  \max_{x^{adv}: \|x-x^{adv}\|_{\infty} \le \epsilon}
  \mathbf{E}_{\theta} \left[\frac{1}{k}\sum_{i=1}^{k} 
    J\left(f^{i}(T_{\theta}(x)), y_{true}\right) \right]\,.
  \]
However, in a typical attack scenario, the true label
$y_{true}$ is not available to the attacker, therefore we substitute
it with a psuedo-label $\hat{y}$ generated by an image
classifer $g$ that is available to the attacker. 
The objective of our attack is thus the following
\[
  \max_{x^{adv}: \|x-x^{adv}\|_{\infty} \le \epsilon}
  \frac{1}{k}\sum_{i=1}^{k} \mathbf{E}_{\theta^{i}} \left[
    J\left(f^{i}(T_{\theta^{i}}(x)), g(x)\right) \right]\,.
    \label{eq:adv_comp:sangxia:obj}
  \]

Using linearity of gradients, we write the gradient of the objective as
\[
  \frac{1}{k}\sum_{i=1}^{k} \nabla_x \mathbf{E}_{\theta^{i}} \left[
    J\left(f^{i}(T_{\theta^{i}}(x)), g(x)\right) \right]\,.
\]
For typical distributions of $\theta$, such as uniform or normal distribution,
the gradient of the expected cross entropy loss over a random $\theta$
is hard to compute. In our solution, we replace it with an empirical estimate
which is an average of the gradients for a few samples of $\theta$.
We also adopt the approach in ~\cite{Tramer2017-EAT} where $x$ is first
randomly perturbed. The use of random projective transformation seems
to be a natural idea, but to the best of our knowledge, this has not
been explicitly described in previous works on generating adversarial 
examples for image classifiers.

In the rest of this section, we use $\widehat{\nabla^{i}}(x)$ to denote
the empirical gradient estimate on input image $x$ as described above.

Let $x^{0}_{adv} := x$, $x^{min} = max(x-\epsilon,0)$, $x^{max}=min(x+\epsilon,1)$,
and let $\alpha^{1}, \alpha^{2}, \ldots$ be a sequence of pre-defined step sizes.
Then in the $i$-th step of the iteration, we update the image by
\[
  x^{i}_{adv} = clip\left(x^{i-1}_{adv} + \alpha^{i} sign\left(
  \frac{1}{k} \sum_{i=1}^{k} \widehat{\nabla^{i}}(x)
  \right), x^{min}, x^{max}\right)\,.
\]

\runinhead{Optimization}
We noticed from our experiments that non-targeted attacks against
pre-trained networks without defense (white-box and black-box)
typically succeed in 3 -- 4 rounds,
whereas attacks against adversarially trained networks take more iterations.
We also observed that in later iterations, there is little benefit in
including in the ensemble un-defended networks that have been 
successfully attacked.
In the final solution, each iteration is defined by step size $\alpha^{i}$
as well as the set of classifiers to include in the ensemble for the
respective iteration. These parameters were found through trial and error
on the official development dataset of the competition.

\runinhead{Experiments: non-targeted attack}
We randomly selected 18,000 images from ImageNet ~\cite{ImageNetChallenge2015}
for which Inception V3 ~\cite{szegedy2015inceptionv3} classified correctly.

The classifiers in the ensemble are: 
Inception V3 ~\cite{szegedy2015inceptionv3}, 
ResNet 50 ~\cite{he2015resnetv1}, ResNet 101 ~\cite{he2015resnetv1},
Inception ResNet V2 ~\cite{szegedy2017inception}, Xception ~\cite{chollet2016xception},
ensemble adversarially trained Inception ResNet V2 
(EnsAdv Inception ResNet V2) ~\cite{Tramer2017-EAT},
and adversarially trained Inception V3 (Adv Inception V3) 
~\cite{Kurakin-AdversarialMlAtScale}.

We held out a few models to evaluate the transferrability of our attacks.
The holdout models listed in Table ~\ref{tab:adv_comp:sangxia:non_targeted}
are: Inception V4 ~\cite{szegedy2017inception},
ensemble adversarially trained Inception V3 with 2 (and 3) 
external models (Ens-3-Adv Inception V3, and Ens-4-Adv Inception V3, 
respectively) ~\cite{Tramer2017-EAT}.

\begin{table}[h]
  \centering
  \caption{Success rate --- non-targeted attack}
  \label{tab:adv_comp:sangxia:non_targeted}
  \begin{tabular}{p{4cm} r}
    \hline
    Classifier & Success rate \\
    \hline
    Inception V3 &  $96.74\%$ \\
    ResNet 50 & $92.78\%$ \\
    Inception ResNet V2 & $92.32\%$ \\
    EnsAdv Inception ResNet V2 & $87.36\%$ \\
    Adv Inception V3 & $83.73\%$ \\
    \hline 
    Inception V4 &  $91.69\%$ \\
    Ens-3-Adv Inception V3 & $62.76\%$ \\
    Ens-4-Adv Inception V3 &  $58.11\%$ \\
    \hline
  \end{tabular}
\end{table}

Table ~\ref{tab:adv_comp:sangxia:non_targeted} lists the success rate
for non-targeted attacks with $\epsilon=16/255$. The performance for
$\epsilon=12/255$ is similar, and somewhat worse for smaller $\epsilon$.
We see that a decent
amount of the generated attacks transfer to the two holdout adversarially 
trained network Ens-3-Adv Inception V3 and Ens-4-Adv Inception V3. 
The transfer rate for many other publicly available pretrained networks 
without defense are all close to or above $90\%$. 
For brevity, we only list the performance on Inception V4 
for comparison.

\runinhead{Targeted attack}
Our targeted attack follows a similar approach as non-targeted attack.
The main differences are:
\begin{enumerate}
  \item For the objective, we now \emph{minimize} the loss between
    a target label $y_{target}$, instead of maximizing with respect to
    $\hat{y}$ as in Equation (\ref{eq:adv_comp:sangxia:obj}).
  \item Our experiments show that doing random image augmentation severely
    decreases the success rate for even \emph{white-box} attacks, therefore
    no augmentation is performed for targeted-attacks. Note that here
    success is defined as successfully make the classifier output the target
    class. The attacks with image augmentation typically managed to
    cause the classifiers to output some wrong label other than the 
    target class.
\end{enumerate}
Our conclusion is that 
if the success criteria is to trick the classifier 
into outputting some specific target class, then
our targeted attack does not transfer well and
is not robust.
 % 2nd place both attacks - team Sangxia
%%\input{submission_defense_3_anlthms.tex} % 3rd place defense - team Anil Thomas
%%\input{submission_nontarget_3_stanford.tex} % 3rd place non-targeted - team Stanford & Sun
%%%%%%%%%%%%%%%%%%%%%%%%%%%%%%%%%%%%%%%%%%%%%%%%%%%%%%%%%%%%%%%%%%%
%
% Subsection about submission XXXXXXXX
%
%%%%%%%%%%%%%%%%%%%%%%%%%%%%%%%%%%%%%%%%%%%%%%%%%%%%%%%%%%%%%%%%%%%

\subsection{3rd place in targeted attack track: team FatFingers}
%\subsection{Dynamic Loss Ensemble Targeted Adversarial Attacks}
\label{sec:adv_comp:submission_t3}

\runinhead{Team members:} Yao Zhao, Yuzhe Zhao, Zhonglin Han and Junjiajia Long

%% Deep neural networks in many tasks such as image classification, are known to be vulnerable to small input perturbations that are unnoticeable by human 1. We here study the targeted adversarial attack scenario, in which the goal is to perturb an image so it will be wrongly classified as a desired target class by some known or unknown classifiers. Several common attack methods exist such as fast gradient sign method (FGSM) [2] and its variants and extensions, optimization-based methods 3,4, Jacobian-based saliency map method (JSMA) 5, and so on. 

We propose a dynamic iterative ensemble targeted attack method, which builds iterative attacks on a loss ensemble neural networks focusing on the classifiers that are harder to perturb. Our methods are tested among 65 attackers against 107 defenders in NIPS-Kaggle competition and achieved 3rd in the targeted attack ranking.

%%\subsubsection{Method details}
%%\label{sec:adv_comp:submission_t3:method}

\subsubsection{Targeted Attack Model Transfer}
In our experiments, we compared variants of single step attack methods and iterative attack methods including two basic forms of those two attack methods: 
fast gradient sign (FGS) 

\begin{equation}\label{eq:adv_comp:submission_t3:fgsm}
\textbf{x}^{adv} =\textbf{x} + \epsilon \cdot sign \bigl( \nabla_x J(f(\textbf{x}), y_{true}) \bigr)
\end{equation}
and iterative sign attacks:

\begin{equation}\label{eq:adv_comp:submission_t3:isa}
\textbf{x}^{adv}_{t+1} = {clip}_{\epsilon, \textbf{x}} \left \{ \textbf{x}^{adv}_t + \alpha \cdot sign \bigl( \nabla_x J(f(\textbf{x}^{adv}_t), y_{true}) \bigr)\right \}
\end{equation}

\begin{figure}[h]

\begin{center}
    \includegraphics[width=\textwidth]{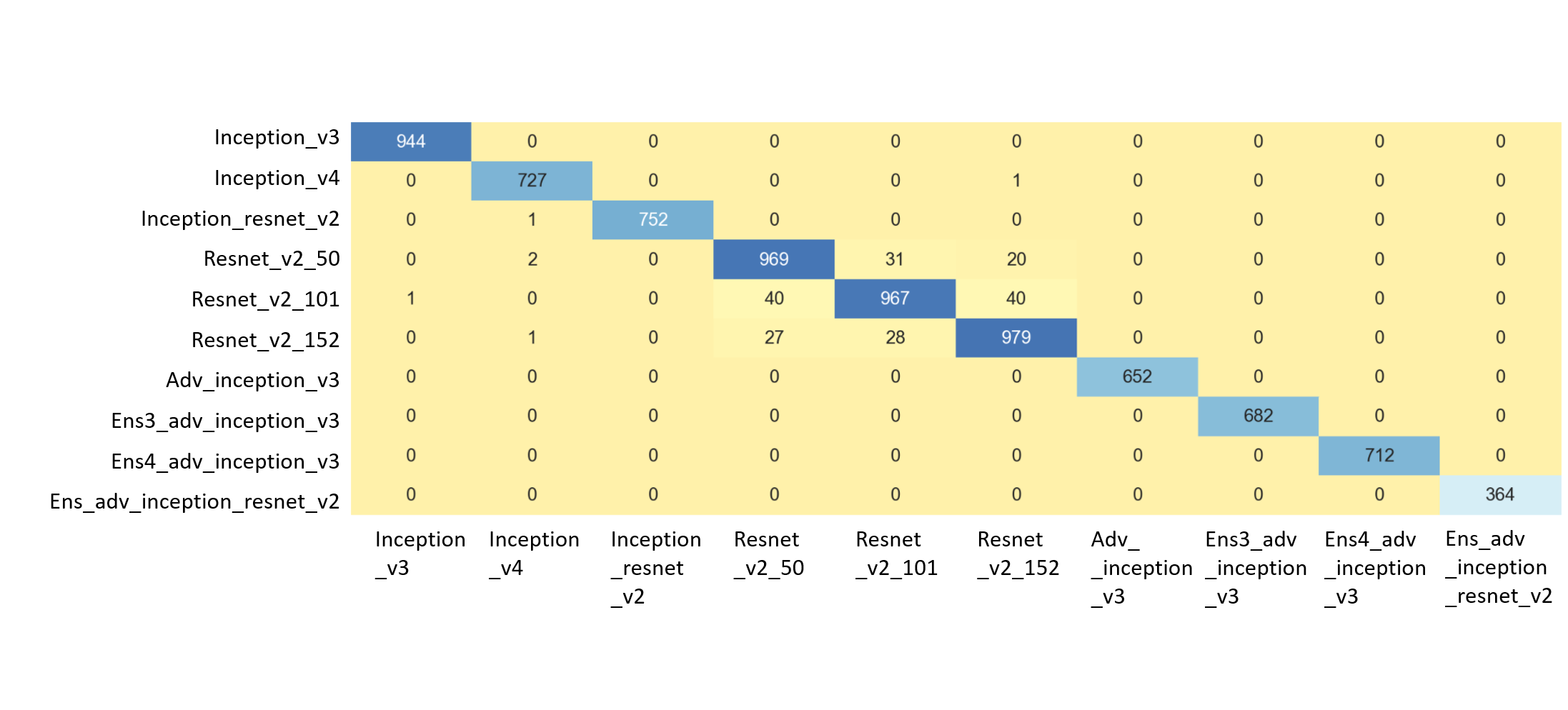}
\end{center}
\caption{Target Hitting Matrix}
\label{fig:adv_comp:submission_t3:hittarget}       % Give a unique label
\end{figure}

\begin{figure}[h]

\begin{center}
    \includegraphics[width=\textwidth]{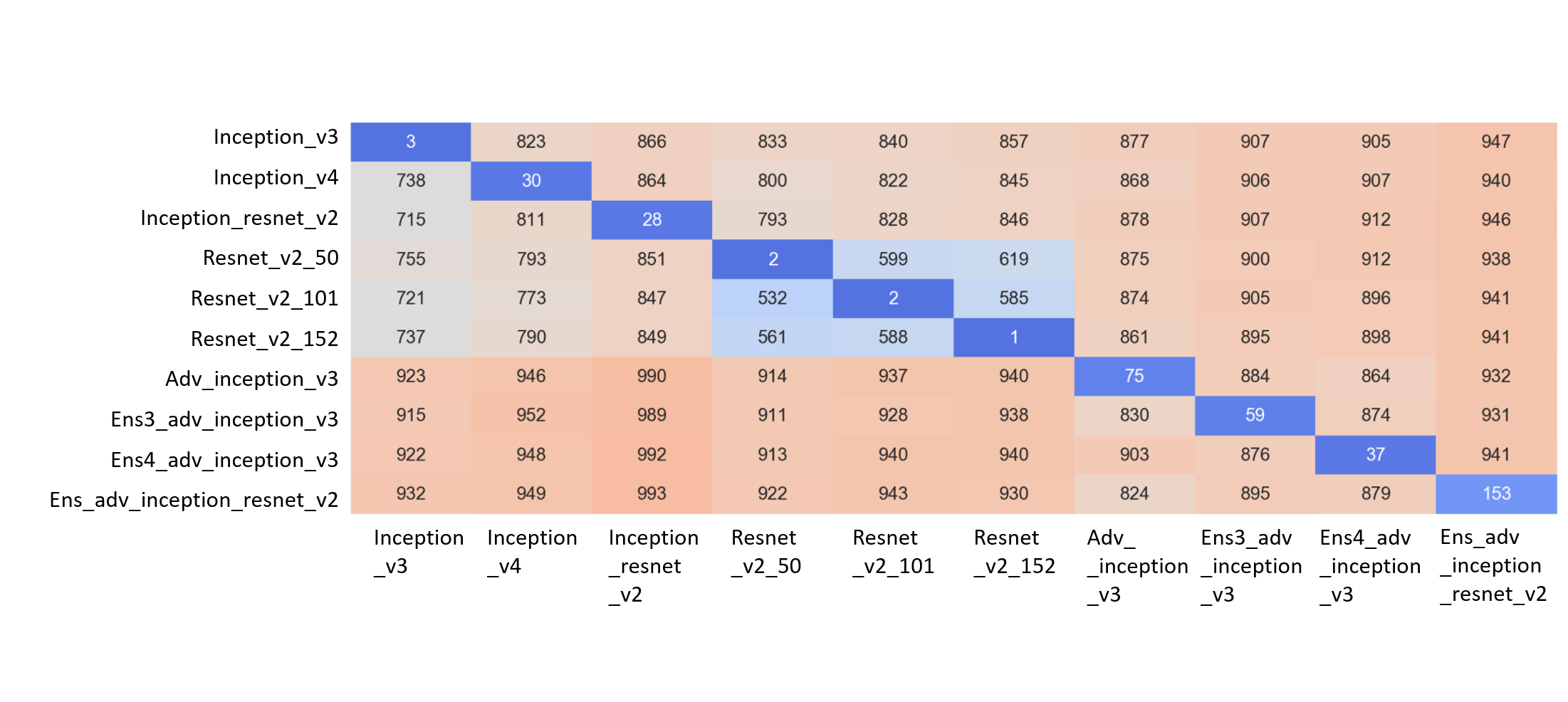}
\end{center}
\caption{Defender Accuracy Matrix}
\label{fig:adv_comp:submission_t3:defense}       % Give a unique label
\end{figure}

To evaluate the ability of black-box targeted attacks, we built iterative attack methods (10 iterations) using single models against many single model defenders individually on 1000 images. Fig.\ref{fig:adv_comp:submission_t3:hittarget} demonstrates the matrix of target hitting for 10 attacking models, while Fig.\ref{fig:adv_comp:submission_t3:defense} shows their capabilitis of defending.

White-box targeted adversarial attacks are generally successful, even against adversarial trained models. Though targeted adversarial attacks built on single models lower the accuracy of defenders based on a different model, the hit rate are close to zero. 

\subsubsection{Ensemble Attack Methods}

Since targeted attacks against unknown models has very low hit rate, it is important to combine known models in a larger number and more efficiently to attack a pool of unknown models or their ensembles.

Probability ensemble is a common way to combine a number of classifiers (sometimes called majority vote). However, the loss function is usually hard to optimize because the parameters of different classifiers are coupled inside the logarithm. 

\begin{equation} \label{eq:adv_comp:submission_t3:prob_ensemble}
J_{prob}\left ( \textbf{x}, y \right ) = -\sum \limits_{j}^N y_j \log\left ( \frac{1}{M} \sum \limits_{i}^M p_{ij} \left ( \textbf{x} \right ) \right ) 
\end{equation}

By Jensen's inequality, an upper bound is obtained for the loss function. Instead of minimizing $J_{prob} (\textbf{x}, y)$, we propose to optimize the upper bound. This way of combining classifiers is called loss ensemble. By using the following new loss function eq.4, the parameters of different neural networks are decoupled, which helps the optimization. 

\begin{equation} \label{eq:adv_comp:submission_t3:loss_ensemble} 
J_{prob}\left ( \textbf{x}, y \right ) \le - \frac{1}{M} \sum \limits_{j}^N \sum \limits_{i}^M y_{ij} \log \left ( p_{ij} \left (\textbf{x} \right ) \right ) = J_{loss} \left ( \textbf{x}, y \right )
\end{equation}

\begin{figure}[h]

\begin{center}
    \includegraphics[width=\textwidth]{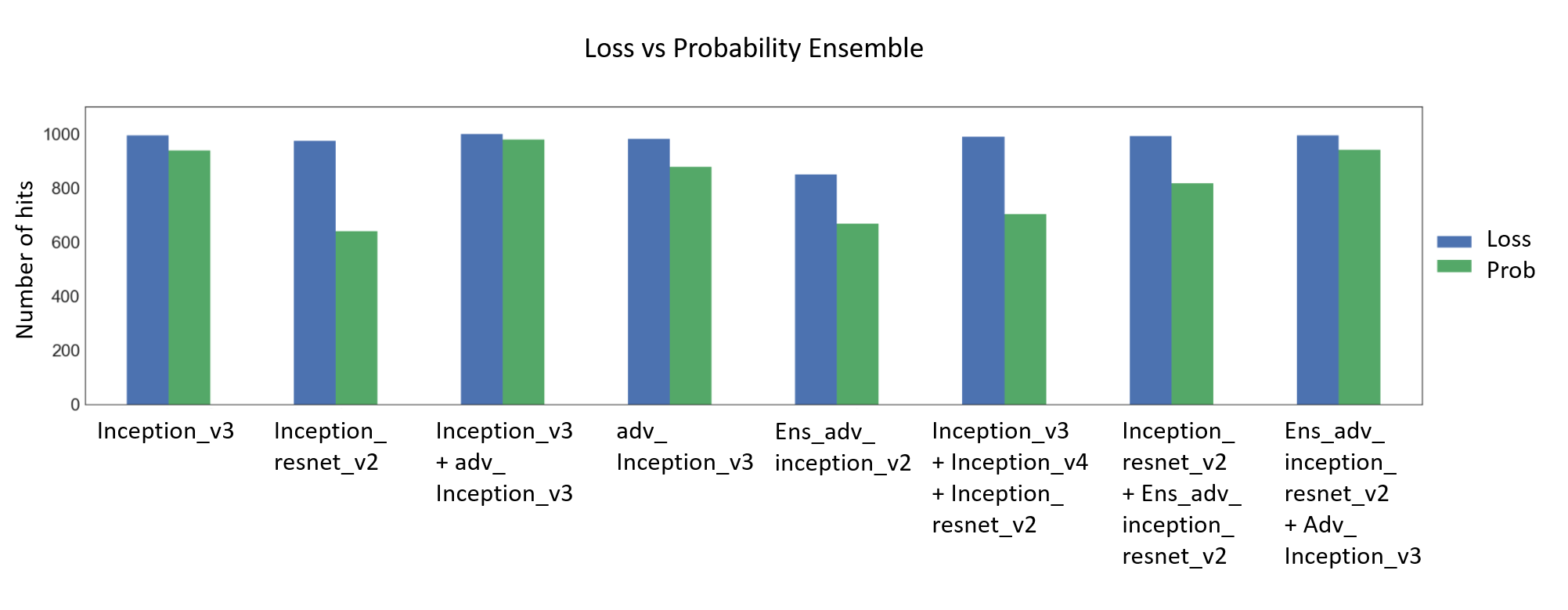}
\end{center}
\caption{Loss ensemble v.s. probability ensemble.
Targeted attacks using the loss ensemble method outperforms probability ensemble at given number of iterations.}
\label{fig:adv_comp:submission_t3:loss_prob}       % Give a unique label
\end{figure}

Comparisons between results of targeted attacks using loss ensemble and probability ensemble at given iterations were shown in Fig.\ref{fig:adv_comp:submission_t3:loss_prob}. In general, it demonstrates that capability of targeted attacking using loss ensemble is superior to that using probability ensemble. 

\subsubsection{Dynamic Iterative Ensemble Attack}

\begin{figure}[h]

\begin{center}
    \includegraphics[width=\textwidth]{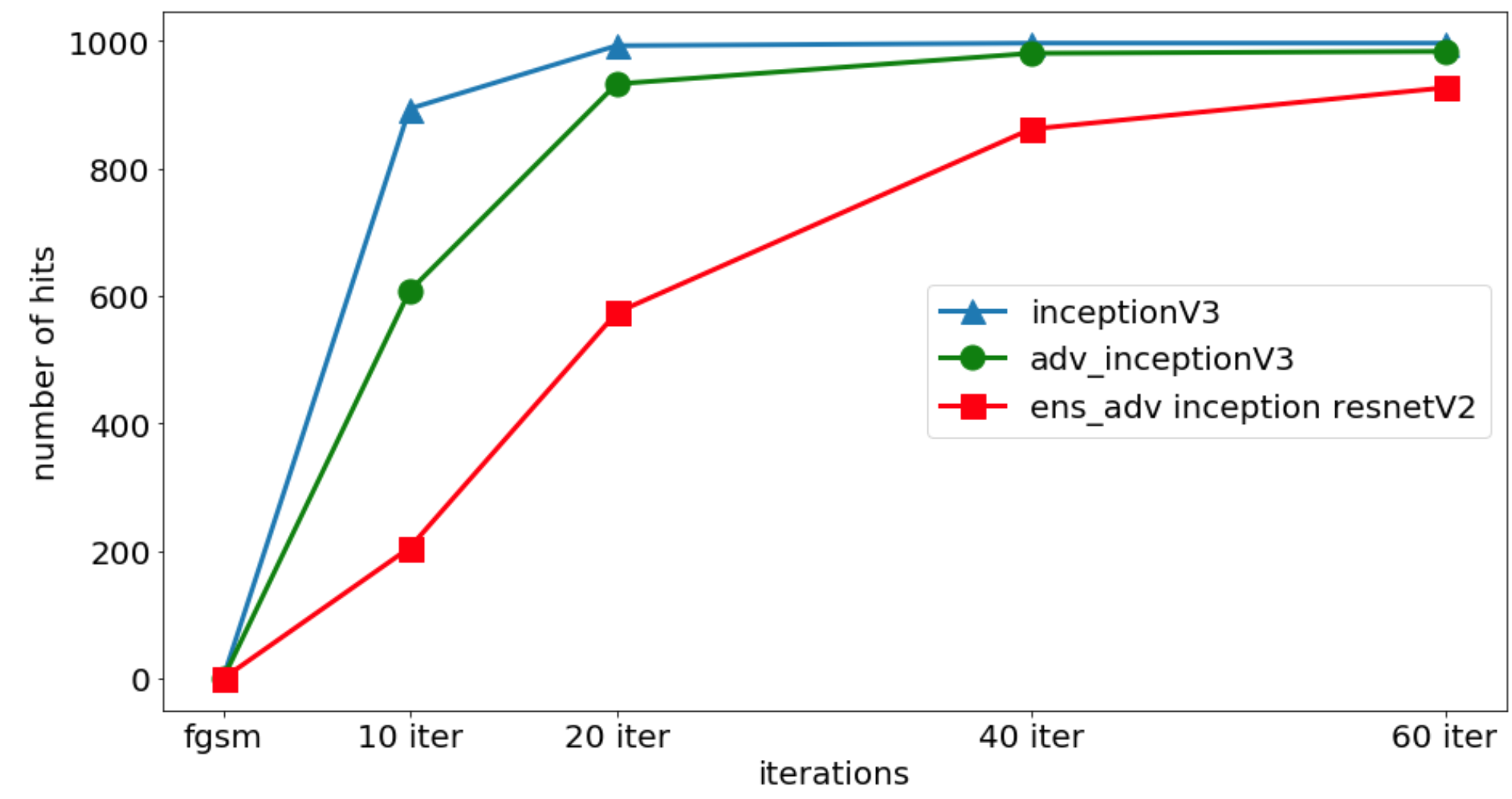}
\end{center}
\caption{Dynamic iterative ensemble attack results for three selected models}
\label{fig:adv_comp:submission_t3:diea}       % Give a unique label
\end{figure}

The difficulty of attacking each individual neural network model within an ensemble can be quite different. We compared iterative attack methods with different parameters and found that number of iterations is most crucial, as shown in Fig.\ref{fig:adv_comp:submission_t3:diea} . For example, attacking an adversarial trained model at high success rate takes significantly more iterations than normal models. 

%%% todo: add one comment for equation 5

\begin{equation}\label{eq:adv_comp:submission_t3:diea}
\textbf{x}^{adv}_{t+1} = {clip}_{\epsilon, \textbf{x}} \left \{ \textbf{x}^{adv}_t + \alpha \cdot sign \bigl( \frac{1}{M}  \sum \limits_{k}^M \delta_{tk} \nabla_x J_k(f(\textbf{x}^{adv}_t), y_{true}) \bigr)\right \}
\end{equation}

For tasks where computation is limited, we implemented a method that pre-assigns
the number of iterations for each model or dynamically adjusts whether to
include a model in each step of the attack by observing if the loss function
for that model is small enough. As shown in Eq.\ref{eq:adv_comp:submission_t3:diea}, $\delta_{tk} \in \{0, 1\}$ determines if loss for model $k$ is included in the total loss at time step $t$.

 % 3rd place targeted - team FatFingers
%%%%%%%%%%%%%%%%%%%%%%%%%%%%%%%%%%%%%%%%%%%%%%%%%%%%%%%%%%%%%%%%%%%
%
% Subsection about submission XXXXXXXX
%
%%%%%%%%%%%%%%%%%%%%%%%%%%%%%%%%%%%%%%%%%%%%%%%%%%%%%%%%%%%%%%%%%%%

\subsection{4th place in defense track: team erko}
\label{sec:adv_comp:submission_erko}

\runinhead{Team members:} Yerkebulan Berdibekov

In this section, I describe a very simple defense solution against adversarial
attacks using spatial smoothing on the input of adversarially trained models.
This solution took 4th place in the final round. Using spatially smoothing,
in particularly median filtering with 2 by 2 windows on images and processing
it by only adversarially trained models we can achieve simple and decent
defense against black box attacks.
Additionally this approach can work along with other defense solutions that
use randomizations (data augmentations \& other types of defenses).

Adversarially trained models are models trained on adversarial examples along
with a given original dataset. In the usual procedure for adversarial training,
during the training phase half of each mini-batch of images are replaced with
adversarial examples generated on the model itself (white box attacks).
This can provide robustness against future white-box attacks.
However, like described in ~\cite{Tramer2017-EAT} $gradient masking$ makes
the finding of adversarial examples a challenging task.
Due to this, adversarially trained models cannot guarantee robustness against
black-box attacks.
Many other techniques have been developed to overcome these problems.

\subsubsection{Architecture of Defense Model}
\label{sec:adv_comp:submission_erko:architecture}

Figure~\ref{fig:adv_comp:submission_erko:architecture} below shows the
architecture of my simple defense model: an input image is followed by median
filtering, and then this filtered image is fed to ensemble of adversarially
trained models. The resulting predictions are then averaged. However, like
described in the sections below, many other variations of ensembles and single
models were tested. The best results were achieved using an ensemble of all
adversarially trained models with median filtering.

\begin{figure}[h]
\centering
% For example, with the graphicx style use
\includegraphics[scale=.25]{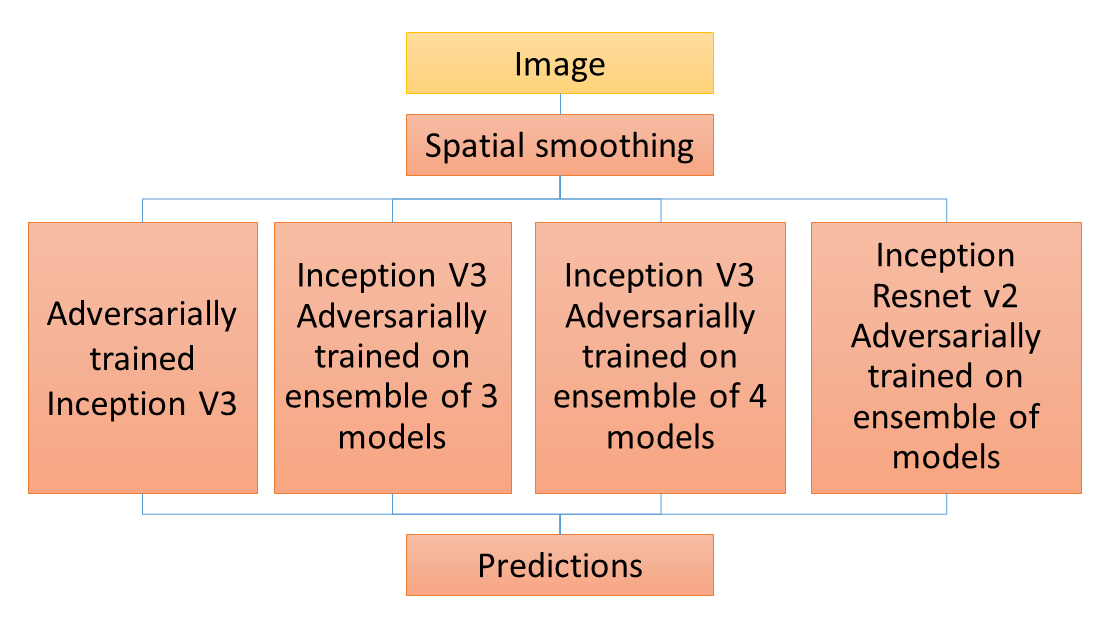}
%
% If no graphics program available, insert a blank space i.e. use
%\picplace{5cm}{2cm} % Give the correct figure height and width in cm
%
\caption{Architecture of simple defense model, using median filtering with
only adversarially trained models.}
\label{fig:adv_comp:submission_erko:architecture}       % Give a unique label
\end{figure}

\subsubsection{Spatial smoothing: median filtering.}
\label{sec:adv_comp:submission_erko:spatial}

Median filtering is often used in image/photo pre-processing to reduce noise
while preserving edges and other features.
It is robust against random high-magnitude perturbations resembling
salt-and-pepper noise. Photographers also use median filtering to increase
photo quality. ImageNet may contain many median filtered images.
Other major advantages of image filtering include:

\begin{itemize}
  \item{Median filtering does not harm classification accuracy on clean
    examples, as shown below in experiments in Section~\ref{sec:adv_comp:submission_erko:experiments}}
  \item{Does not require additional expensive training procedures other than
    the adversarially trained model itself.}
\end{itemize}

\subsubsection{Experiments}
\label{sec:adv_comp:submission_erko:experiments}

I have experimentally observed that using median filtering only we cannot
defend against strong adversarial attacks like described by Carlini and
Wagner ~\cite{Carlin-Wagner-attack}.
However, I have also experimentally observed that using median filtering
and only adversarially trained models we can obtain a robust defense against
adversarial attacks.

In my experiments I used the dataset provided by competition organizers and
used a modified C\&W L2 attack to generate adversarial examples.
These examples were later used to calculate the adversarial example
misclassification ratio (number of wrong classifications divided by number
of all examples) and to rank defenses. To generate adversarial examples I
used either a single model or ensemble of models (a list of multiple models
is indicated in each cell).

In all experiments I used a hold-out \verb|inception_v4| model that was not
used to generate adversarial examples (see
Table~\ref{tab:adv_comp:submission_erko:table1},
Table~\ref{tab:adv_comp:submission_erko:table2}).
This allowed us to test transferability of attacks and to test spatial
smoothing effects.

\subsubsection{Effects of median filtering}
\label{sec:adv_comp:submission_erko:effects}

On our holdout \verb|inception_v4| model, using median filtering performs nearly the same as without median filtering. Same results on other non-adversarially trained models. With median filtering or without, misclassification ratio differences are small.

Adversarially trained models with median filtering show good defense against attacks. An ensemble of these adversarially trained models with median filtered images is robust against black-box attacks and to attacks generated by an ensemble containing same models (see Table~\ref{tab:adv_comp:submission_erko:table1}, Table~\ref{tab:adv_comp:submission_erko:table2}). This is not exactly a white-box attack, because we generate adversarial examples on a model without a filtering layer. For example, we use a pre-trained \verb|ens3_adv_inception_v3| model to generate adversarial examples. These images are median filtered and fed to model again to check the misclassification ratios. 

All these attacks were generated using $\epsilon$=16 max pixel perturbations. In the case of the best ensemble defense against the best ensemble attacker, I tested other values of $\epsilon$ and plotted Figure~\ref{fig:adv_comp:submission_erko:graph1}, showing that in case of lower $\epsilon$ values this defense approach is more robust against attacks(exact values in Table~\ref{tab:adv_comp:submission_erko:table3}):

\begin{figure}[h]
\centering
% For example, with the graphicx style use
\includegraphics[scale=.4]{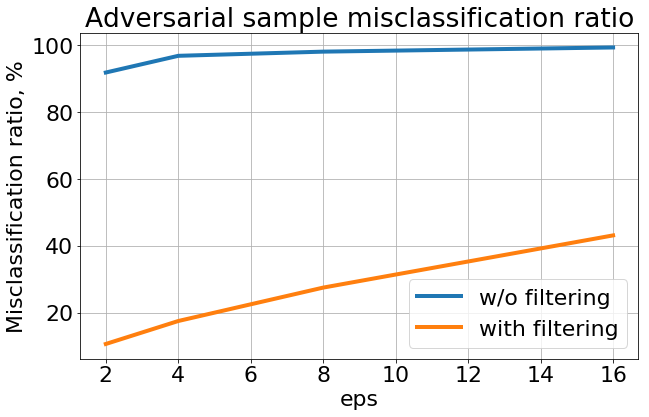}
%
% If no graphics program available, insert a blank space i.e. use
%\picplace{5cm}{2cm} % Give the correct figure height and width in cm
%
\caption{Adversarial examples misclassification ratio, percentage}
\label{fig:adv_comp:submission_erko:graph1}       % Give a unique label
\end{figure}

\newcommand{\specialcell}[2][c]{%
  \begin{tabular}[#1]{@{}l@{}}#2\end{tabular}}

\begin{table}[h]
\centering
\caption{Misclassification ratio without filtering, percentage. Rows are defenders; columns are attackers. Even ensemble of adversarially trained models are not robust against good attackers.}
\label{tab:adv_comp:submission_erko:table1}       % Give a unique label
%
% Follow this input for your own table layout
%
\setlength\tabcolsep{6pt}
\begin{tabular}{p{4.9cm}cccc}
\hline\noalign{\smallskip}
Defenders\textbackslash{}Attackers & inception\_v3 & $A$ & $B$ & $C$ \\
\noalign{\smallskip}\svhline\noalign{\smallskip}
inception\_v3 & 100.00 & 100.00 & 26.25 & 99.38\\\hline
inception\_v4 & 42.50 & 80.63 & 21.88 & 62.50\\\hline
adv\_inception\_v3 & 20.62 & 41.25 & 100.00 & 100.00\\\hline
ens3\_adv\_inception\_v3 & 15.62 & 38.13 & 100.00 & 99.38\\\hline
ens\_adv\_inception\_resnet\_v2 & 10.62 & 23.75 & 94.38 & 95.00\\\hline
\specialcell[t]{adv\_inception\_v3\\ens3\_adv\_inception\_v3} & 15.00 & 36.25 & 100.00 & 100.00\\\hline
\specialcell[t]{adv\_inception\_v3\\ens3\_adv\_inception\_v3\\ens4\_adv\_inception\_v3} & 16.25 & 33.13 & 100.00 & 99.38\\\hline
\specialcell[t]{adv\_inception\_v3\\ens3\_adv\_inception\_v3\\ens\_adv\_inception\_resnet\_v2\\ens4\_adv\_inception\_v3} & 12.5 & 28.75 & 100.00 & 99.38\\
\noalign{\smallskip}\hline\noalign{\smallskip}
\end{tabular}
\\
\raggedright
Where $A$ is an ensemble of inception\_v3, inception\_resnet\_v2, resnet\_v1\_101, resnet\_v1\_50, resnet\_v2\_101, resnet\_v2\_50, vgg\_16;\\
$B$ is an ensemble of adv\_inception\_v3, ens3\_adv\_inception\_v3, ens\_adv\_inception\_resnet\_v2, ens4\_adv\_inception\_v3;\\
$C$ is an ensember of inception\_v3, adv\_inception\_v3, ens3\_adv\_inception\_v3, ens\_adv\_inception\_resnet\_v2, ens4\_adv\_inception\_v3, inception\_resnet\_v2, resnet\_v1\_101, resnet\_v1\_50, resnet\_v2\_101.
\end{table}
%

\begin{table}[h]
\centering
\caption{Misclassification ratio with filtering, percentage. Adversarially trained models with median filtering show
 better robustness against many kinds of attacks within these experiments. inception\_v4 model with median filtering on all of attacks performs nearly same as without filtering. Same on other non-adversarial models. Therefore, I am speculating median filtering is not cleaning, or not mitigating adversarial examples.}
\label{tab:adv_comp:submission_erko:table2}       % Give a unique label
%
% Follow this input for your own table layout
%
%\begin{tabular}{p{4.9cm}p{1.5cm}p{1.5cm}p{1.5cm}p{1.5cm}}
\setlength\tabcolsep{6pt}
\begin{tabular}{p{4.9cm}cccc}
\hline\noalign{\smallskip}
Defenders\textbackslash{}Attackers & inception\_v3 & $A$ & $B$ & $C$ \\
\noalign{\smallskip}\svhline\noalign{\smallskip}
inception\_v3 & 100.00 & 97.50 & 27.50 & 95.63\\\hline
inception\_v4 & 40.00 & 75.63 & 22.50 & 57.50\\\hline
adv\_inception\_v3 & 21.88 & 43.13 & 33.13 & 40.00\\\hline
ens3\_adv\_inception\_v3 & 21.88 & 43.75 & 57.50 & 58.13\\\hline
ens\_adv\_inception\_resnet\_v2 & 13.13 & 30.63 & 30.63 & 39.38\\\hline
\specialcell[t]{adv\_inception\_v3\\ens3\_adv\_inception\_v3} & 17.50 & 40.00 & 43.75 & 47.50\\\hline
\specialcell[t]{adv\_inception\_v3\\ens3\_adv\_inception\_v3\\ens4\_adv\_inception\_v3} & 17.50 & 38.75 & 43.75 & 48.75\\\hline
\specialcell[t]{adv\_inception\_v3\\ens3\_adv\_inception\_v3\\ens\_adv\_inception\_resnet\_v2\\ens4\_adv\_inception\_v3} & 14.38 & 35.00 & 39.38 & 43.13\\
\noalign{\smallskip}\hline\noalign{\smallskip}
\end{tabular}
\\
\raggedright
Where $A$ is an ensemble of inception\_v3, inception\_resnet\_v2, resnet\_v1\_101, resnet\_v1\_50, resnet\_v2\_101, resnet\_v2\_50, vgg\_16;\\ 
$B$ is an ensemble of adv\_inception\_v3, ens3\_adv\_inception\_v3, ens\_adv\_inception\_resnet\_v2, ens4\_adv\_inception\_v3;\\ 
$C$ is an ensemble of inception\_v3, adv\_inception\_v3, ens3\_adv\_inception\_v3, ens\_adv\_inception\_resnet\_v2, ens4\_adv\_inception\_v3, inception\_resnet\_v2, resnet\_v1\_101, resnet\_v1\_50, resnet\_v2\_101.
\end{table}
%

\begin{table}[h]
\centering
\caption{Misclassification ratio on $\epsilon$ values, percentage. On smaller $\epsilon$ values, median filtering shows even better robustness to adversarial attacks.}
\label{tab:adv_comp:submission_erko:table3}       % Give a unique label
%
% Follow this input for your own table layout
%
\setlength\tabcolsep{6pt}
%\begin{tabular}{p{4.9cm}cccc}
\begin{tabular}{p{6cm}cccc}
\hline\noalign{\smallskip}
Defenders & $\epsilon$=16 & $\epsilon$=8 & $\epsilon$=4 & $\epsilon$=2  \\
\noalign{\smallskip}\svhline\noalign{\smallskip}
Ensemble of adversarial models non-filtered input & 99.375 & 98.125 & 96.875 & 91.875\\
Ensemble of adversarial models with filtered input & 43.125 & 27.500 & 17.500 & 10.625\\
\noalign{\smallskip}\hline\noalign{\smallskip}
\end{tabular}
\end{table}

\subsubsection{Submission results}
\label{sec:adv_comp:submission_erko:results}

Following the competition results, we have seen that adversarially trained models with median filtering are indeed robust to most types of attacks.
These results suggest more study on this effect of adversarially trained
models in the future.

During the competition, new types of attacks were developed with smoothed
adversarial examples that can fool spatially smoothed defenses with as high
as 50-60\% ratio and with high transferability.
These are the best attackers developed in Non-Targeted/Targeted Adversarial
Attack Competitions.
Additional study is needed to defend against these new types of attacks.

 % 4th place defense - team erko
%%%%%%%%%%%%%%%%%%%%%%%%%%%%%%%%%%%%%%%%%%%%%%%%%%%%%%%%%%%%%%%%%%%
%
% Subsection about submission XXXXXXXX
%
%%%%%%%%%%%%%%%%%%%%%%%%%%%%%%%%%%%%%%%%%%%%%%%%%%%%%%%%%%%%%%%%%%%

% Sorry for making new our own command! We hope the default `paragraph` to be improved!
\newcommand{\myparagraph}[1]{\vspace{0.5em} \noindent {\bf #1.}}

\subsection{4th place in non-targeted attack track: team iwiwi}

\runinhead{Team members:} Takuya Akiba and Seiya Tokui and Motoki Abe

In this section, we explain the submission from team \emph{iwiwi} to the non-targeted attack track.
This team was Takuya Akiba, Seiya Tokui and Motoki Abe.
The approach is quite different from other teams:
training fully-convolutional networks (FCNs) that can convert clean examples to adversarial examples.
The team received the 4th place.

\subsubsection{Basic Framework}
Given a clean input image $x$, we generate an adversarial example as follows:
\begin{equation*}
x^{adv} = {Clip}_{[0, 1]}( x + a(x; \theta_a) ).
\end{equation*}
Here, $a$ is a differentiable function represented by a FCN
with parameter $\theta_a$.
We call $a$ as an \emph{attack FCN}.
It outputs $c \times h \times w$ tensors, where $c, h, w$ are the number of channels, height and width of $x$.
The values of the output are in range $[-\varepsilon, +\varepsilon]$.
During the training of the attack FCN, to confuse image classifiers,
we maximize the loss $J(f(x^{adv}), y)$, where $f$ is a pre-trained image classifier.
We refer to $f$ as a target model.
Specifically, we optimize $\theta_a$ to maximize the following value:
\begin{equation*}
\sum_{x \in \mathcal{X}} J \left(f \left( {Clip}_{[0, 1]} \left( x + a \left(x; \theta_a \right) \right) \right), y \right).
\end{equation*}

This framework has some commonality with the work by Baluja and Fischer~\cite{ATN2017}.
They also propose to train neural networks that produce adversarial examples.
However, while we have the hard constraint on the distance between clean and adversarial examples,
they considered the distance as one of optimization objective to minimize.
In addition, we used a much larger FCN model and stronger computation power, together with several new ideas such as multi-target training, multi-task training, and gradient hints, which are explained in the next subsection.

\subsubsection{Empirical Enhancement}

\myparagraph{Multi-Target Training}
To obtain adversarial examples that generalize to different image classifiers,
we use multiple target models to train the attack FCN.
We maximize the sum of losses of all models.
In this competition, we used eight models:
\textit{(1)} ResNet50,
\textit{(2)} VGG16,
\textit{(3)} Inception v3,
\textit{(4)} Inception v3 with adversarial training,
\textit{(5)} Inception v3 with ensemble adversarial training (EAT) using three models,
\textit{(6)} Inception v3 with EAT using four models,
\textit{(7)} Inception ResNet v2, and
\textit{(8)} Inception ResNet v2 with EAT.
All of these classifier models are available online.

\myparagraph{Multi-Task Training}
A naive approach to construct a FCN so that it outputs values in the range $[-\epsilon, +\epsilon]$
is to apply the $\text{tanh}$ function to the last output, and then multiply it by $\epsilon$.
However, in this way, the FCN cannot finely control the magnitude of perturbation, as $\epsilon$ is not given to the FCN.
To cope with this issue, we take the advantage of discreteness.
In this competition, $\epsilon$ can take 13 values: $\frac{4}{256}, \frac{5}{256}, \ldots, \frac{16}{256}$.
We consider adversarial attack with different $\epsilon$ values as different tasks, and employ multi-task training.
Specifically, the FCN outputs a tensor with shape $13 \times c \times h \times w$,
where the first dimension corresponds to the $\epsilon$ value.

\myparagraph{Gradient Hints}
Attack methods that use the gradients on image pixels work well.
Therefore, these gradients are useful signals for generating adversarial examples.
Thus, in addition to clean examples,
we also use these gradients as input to the FCN.
In this competition, we used gradients by Inception ResNet v2 with EAT,
which was the strongest defense model publicly available.

% \myparagraph{Recurrent Networks}
% Using gradient

% \subsubsection{Training Parallalelization}
% Clever parallelization by combining data and model parallelism.

\subsubsection{Results and Discussion}
% We used 128 GPUs

The team ranked 4th among about one hundred teams.
In addition, the team ranked 1st in 3rd-party PageRank-like analysis\footnote{\url{https://www.kaggle.com/anlthms/pagerank-ish-scoring}},
which shows that this attack method is especially effective for strong defense methods.

\begin{figure}[t]
  \centering
  \includegraphics[width=.32\hsize]{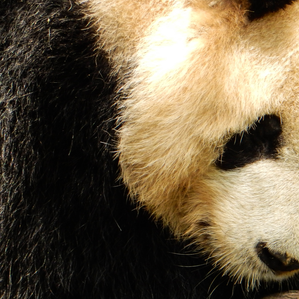}
  \includegraphics[width=.32\hsize]{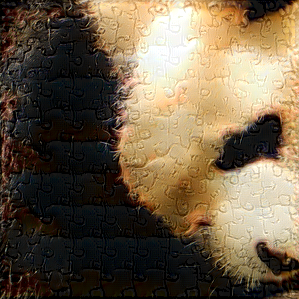}
  \includegraphics[width=.32\hsize]{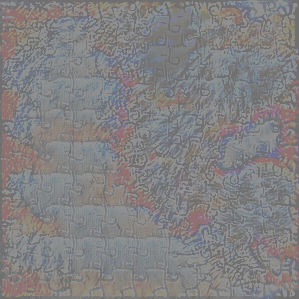}
  \caption{A clean example (left), adversarial example generated by our method (middle), and their difference (right),where $\epsilon = \frac{16}{255}$.}
  \label{fig:iwiwi}
\end{figure}

In addition to its effectiveness, the generated attack images have interesting appearance (Figure~\ref{fig:iwiwi},
more examples are available online\footnote{\url{https://github.com/pfnet-research/nips17-adversarial-attack}}).
We observe two properties from the generated images:
detailed textures are canceled out, and Jigsaw-puzzle-like patterns are added.
These properties deceive image classifiers into answering the Jigsaw puzzle class.

 % 4th place non-targeted - team iwiwi
%%\input{submission_target_4_anlthms.tex} % 4th place targeted - team Anil Thomas

\section{Conclusion}

Adversarial examples are interesting phenomenon and important problem in machine learning security.
Main goals of this competition were to increase awareness of the problem
and stimulate researchers to propose novel approaches.

Competition definitely helped to increase awareness of the problem.
Article ``AI Fight Club Could Help Save Us from a Future of Super-Smart Cyberattacks''\footnote{ \url{www.technologyreview.com/s/608288}}
was published in MIT Technology review about the competition.
And more than 100 teams were competing in the final round.

Competition also pushed people to explore new approaches and improve existing methods to the problem.
In all three tracks, competitors showed significant improvements on top of provided baselines by the end of the competition.
Additionally, top submission in the defense tracked showed $95\%$ accuracy on all adversarial images produced by all attacks.
While worst case accuracy was not as good as an average accuracy,
the results are still suggesting that practical applications may be able to achieve reasonable level of robustness to adversarial
examples in black box case.

\bibliographystyle{abbrv}
\bibliography{adversarial_competition,ml}

\begin{thebibliography}{10}

\bibitem{ATN2017}
S.~Baluja and I.~Fischer.
\newblock Adversarial transformation networks: Learning to generate adversarial
  examples.
\newblock 2017.

\bibitem{biggio2013evasion}
B.~Biggio, I.~Corona, D.~Maiorca, B.~Nelson, N.~{\v{S}}rndi{\'c}, P.~Laskov,
  G.~Giacinto, and F.~Roli.
\newblock Evasion attacks against machine learning at test time.
\newblock In {\em Joint European Conference on Machine Learning and Knowledge
  Discovery in Databases}, pages 387--402. Springer, 2013.

\bibitem{Brendel2017-DecisionBasedBlackBox}
W.~Brendel, J.~Rauber, and M.~Bethge.
\newblock Decision-based adversarial attacks: Reliable attacks against
  black-box machine learning models.
\newblock 2017.

\bibitem{thermometer_enconding2018}
J.~Buckman, A.~Roy, C.~Raffel, and I.~Goodfellow.
\newblock Thermometer encoding: One hot way to resist adversarial examples.
\newblock {\em Submissions to International Conference on Learning
  Representations}, 2018.

\bibitem{Carlini2017-Breaking10Detectors}
N.~Carlini and D.~Wagner.
\newblock Adversarial examples are not easily detected: Bypassing ten detection
  methods.
\newblock In {\em USENIX Workshop on Offensive Technologies}, 2017.

\bibitem{CarliniWagnerAttack}
N.~Carlini and D.~Wagner.
\newblock Towards evaluating the robustness of neural networks.
\newblock {\em IEEE Symposium on Security and Privacy}, 2017.

\bibitem{Zoo2017-ZerothOrderBlackBox}
P.-Y. Chen, H.~Zhang, Y.~Sharma, J.~Yi, and C.-J. Hsieh.
\newblock Zoo: Zeroth order optimization based black-box attacks to deep neural
  networks without training substitute models.
\newblock 2017.

\bibitem{chollet2016xception}
F.~Chollet.
\newblock Xception: Deep learning with depthwise separable convolutions, 2016.

\bibitem{das2017JpegDefense}
N.~Das, M.~Shanbhogue, S.-T. Chen, F.~Hohman, L.~Chen, M.~E. Kounavis, and
  D.~H. Chau.
\newblock Keeping the bad guys out: Protecting and vaccinating deep learning
  with jpeg compression.
\newblock {\em arXiv preprint arXiv:1705.02900}, 2017.

\bibitem{deng2009imagenet}
J.~Deng, W.~Dong, R.~Socher, L.-J. Li, K.~Li, and L.~Fei-Fei.
\newblock Imagenet: A large-scale hierarchical image database.
\newblock In {\em Computer Vision and Pattern Recognition, 2009. CVPR 2009.
  IEEE Conference on}, pages 248--255. IEEE, 2009.

\bibitem{dong2017boosting}
Y.~Dong, F.~Liao, T.~Pang, H.~Su, X.~Hu, J.~Li, and J.~Zhu.
\newblock Boosting adversarial attacks with momentum.
\newblock {\em arXiv preprint arXiv:1710.06081}, 2017.

\bibitem{Korczak1998Optimization}
W.~Duch and J.~Korczak.
\newblock Optimization and global minimization methods suitable for neural
  networks.
\newblock {\em Neural computing surveys}, 2:163--212, 1998.

\bibitem{hinton2018matrix}
N.~F. Geoffrey E~Hinton, Sara~Sabour.
\newblock Matrix capsules with em routing.
\newblock In {\em International Conference on Learning Representations}, 2018.

\bibitem{adversarial_sphere2018}
J.~Gilmer, L.~Metz, F.~Faghri, S.~S. Schoenholz, M.~Raghu, M.~Wattenberg, and
  I.~Goodfellow.
\newblock Adversarial spheres.
\newblock {\em Submissions to International Conference on Learning
  Representations}, 2018.

\bibitem{Goodfellow-2015-adversarial}
I.~J. Goodfellow, J.~Shlens, and C.~Szegedy.
\newblock Explaining and harnessing adversarial examples.
\newblock {\em CoRR}, abs/1412.6572, 2014.

\bibitem{goodfellow2014explaining}
I.~J. Goodfellow, J.~Shlens, and C.~Szegedy.
\newblock Explaining and harnessing adversarial examples.
\newblock {\em CoRR}, abs/1412.6572, 2014.

\bibitem{he2015resnetv1}
K.~He, X.~Zhang, S.~Ren, and J.~Sun.
\newblock Deep residual learning for image recognition, 2015.

\bibitem{he2016identity}
K.~He, X.~Zhang, S.~Ren, and J.~Sun.
\newblock Identity mappings in deep residual networks.
\newblock In {\em ECCV}, 2016.

\bibitem{Warren2017-BreakingEnsembleWeakDefenses}
W.~He, J.~Wei, X.~Chen, N.~Carlini, and D.~Song.
\newblock Adversarial example defense: Ensembles of weak defenses are not
  strong.
\newblock In {\em 11th {USENIX} Workshop on Offensive Technologies ({WOOT}
  17)}, Vancouver, BC, 2017. {USENIX} Association.

\bibitem{LearningWithStrongAdversary}
R.~Huang, B.~Xu, D.~Schuurmans, and C.~Szepesv{\'{a}}ri.
\newblock Learning with a strong adversary.
\newblock {\em CoRR}, abs/1511.03034, 2015.

\bibitem{kingma2014adam}
D.~Kingma and J.~Ba.
\newblock Adam: A method for stochastic optimization.
\newblock {\em arXiv preprint arXiv:1412.6980}, 2014.

\bibitem{PhysicalAdversarialExamples}
A.~Kurakin, I.~Goodfellow, and S.~Bengio.
\newblock Adversarial examples in the physical world.
\newblock In {\em ICLR'2017 Workshop}, 2016.

\bibitem{Kurakin-PhysicalAdversarialExamples}
A.~Kurakin, I.~Goodfellow, and S.~Bengio.
\newblock Adversarial examples in the physical world.
\newblock In {\em ICLR'2017 Workshop}, 2016.

\bibitem{Kurakin-AdversarialMlAtScale}
A.~Kurakin, I.~J. Goodfellow, and S.~Bengio.
\newblock Adversarial machine learning at scale.
\newblock In {\em ICLR'2017}, 2016.

\bibitem{liao2017defense}
F.~Liao, M.~Liang, Y.~Dong, T.~Pang, J.~Zhu, and X.~Hu.
\newblock Defense against adversarial attacks using high-level representation
  guided denoiser.
\newblock {\em arXiv preprint arXiv:1712.02976}, 2017.

\bibitem{liu2017delving}
Y.~Liu, X.~Chen, C.~Liu, and D.~Song.
\newblock Delving into transferable adversarial examples and black-box attacks.
\newblock In {\em Proceedings of 5th International Conference on Learning
  Representations}, 2017.

\bibitem{MadryPgd2017}
A.~Madry, A.~Makelov, L.~Schmidt, D.~Tsipras, and A.~Vladu.
\newblock Towards deep learning models resistant to adversarial attacks.
\newblock 2017.

\bibitem{metzen2017detecting}
J.~H. Metzen, T.~Genewein, V.~Fischer, and B.~Bischoff.
\newblock On detecting adversarial perturbations.
\newblock In {\em ICLR}, 2017.

\bibitem{Papernot-2016-TransferabilityStudy}
N.~{Papernot}, P.~{McDaniel}, and I.~{Goodfellow}.
\newblock {Transferability in Machine Learning: from Phenomena to Black-Box
  Attacks using Adversarial Samples}.
\newblock {\em ArXiv e-prints}, May 2016b.

\bibitem{Papernot-2016-IntroTransferability}
N.~Papernot, P.~McDaniel, I.~Goodfellow, S.~Jha, Z.~B. Celik, and A.~Swami.
\newblock Practical black-box attacks against machine learning.
\newblock In {\em Proceedings of the 2017 ACM on Asia Conference on Computer
  and Communications Security}, ASIA CCS '17, pages 506--519, New York, NY,
  USA, 2017. ACM.

\bibitem{Poljak1964Some}
B.~T. Polyak.
\newblock Some methods of speeding up the convergence of iteration methods.
\newblock {\em USSR Computational Mathematics and Mathematical Physics},
  4(5):1--17, 1964.

\bibitem{ImageNetChallenge2015}
O.~Russakovsky, J.~Deng, H.~Su, J.~Krause, S.~Satheesh, S.~Ma, Z.~Huang,
  A.~Karpathy, A.~Khosla, M.~Bernstein, A.~C. Berg, and L.~Fei-Fei.
\newblock Imagenet large scale visual recognition challenge.
\newblock {\em International Journal of Computer Vision}, 115(3):211--252, Dec
  2015.

\bibitem{Sharif16AdvML}
M.~Sharif, S.~Bhagavatula, L.~Bauer, and M.~K. Reiter.
\newblock Accessorize to a crime: {R}eal and stealthy attacks on
  state-of-the-art face recognition.
\newblock In {\em Proceedings of the 23rd ACM SIGSAC Conference on Computer and
  Communications Security}, Oct. 2016.
\newblock To appear.

\bibitem{Sutskever2013On}
I.~Sutskever, J.~Martens, G.~Dahl, and G.~Hinton.
\newblock On the importance of initialization and momentum in deep learning.
\newblock In {\em ICML}, 2013.

\bibitem{szegedy2017inception}
C.~Szegedy, S.~Ioffe, V.~Vanhoucke, and A.~A. Alemi.
\newblock Inception-v4, inception-resnet and the impact of residual connections
  on learning.
\newblock In {\em AAAI}, 2017.

\bibitem{szegedy2015inceptionv3}
C.~Szegedy, V.~Vanhoucke, S.~Ioffe, J.~Shlens, and Z.~Wojna.
\newblock Rethinking the inception architecture for computer vision, 2015.

\bibitem{Szegedy2016Rethinking}
C.~Szegedy, V.~Vanhoucke, S.~Ioffe, J.~Shlens, and Z.~Wojna.
\newblock Rethinking the inception architecture for computer vision.
\newblock In {\em CVPR}, 2016.

\bibitem{szegedy2104intriguing}
C.~Szegedy, W.~Zaremba, I.~Sutskever, J.~Bruna, D.~Erhan, I.~Goodfellow, and
  R.~Fergus.
\newblock Intriguing properties of neural networks.
\newblock In {\em International Conference on Learning Representations}, 2014.

\bibitem{Szegedy-ICLR2014}
C.~Szegedy, W.~Zaremba, I.~Sutskever, J.~Bruna, D.~Erhan, I.~J. Goodfellow, and
  R.~Fergus.
\newblock Intriguing properties of neural networks.
\newblock {\em ICLR}, abs/1312.6199, 2014.

\bibitem{Tramer2017-EAT}
F.~Tramèr, A.~Kurakin, N.~Papernot, I.~Goodfellow, D.~Boneh, and P.~McDaniel.
\newblock Ensemble adversarial training: Attacks and defenses.
\newblock In {\em arxiv}, 2017.

\bibitem{vincent2008extracting}
P.~Vincent, H.~Larochelle, Y.~Bengio, and P.-A. Manzagol.
\newblock Extracting and composing robust features with denoising autoencoders.
\newblock In {\em International Conference on Machine learning}, pages
  1096--1103, 2008.

\bibitem{xie2017mitigating}
C.~Xie, J.~Wang, Z.~Zhang, Z.~Ren, and A.~Yuille.
\newblock Mitigating adversarial effects through randomization.
\newblock In {\em International Conference on Learning Representations}, 2018.

\bibitem{Weilin2017-FeatureSqueezing}
W.~Xu, D.~Evans, and Y.~Qi.
\newblock Feature squeezing: Detecting adversarial examples in deep neural
  networks.
\newblock {\em CoRR}, abs/1704.01155, 2017.

\bibitem{zhang2017beyond}
K.~Zhang, W.~Zuo, Y.~Chen, D.~Meng, and L.~Zhang.
\newblock Beyond a gaussian denoiser: Residual learning of deep cnn for image
  denoising.
\newblock {\em IEEE Transactions on Image Processing}, 2017.

\end{thebibliography}

\end{document}